\documentclass[11pt, letterpaper]{arxiv}
\usepackage[all]{hypcap}
\usepackage[comma,sort,compress]{natbib}
\usepackage{hyperref}[citecolor=magenta]
\usepackage[capitalize,noabbrev]{cleveref}
\hypersetup{
    colorlinks = true,
    citecolor = {LightBlue},
    linkcolor = {BrickRed}
}

\usepackage{microtype}

\usepackage{subcaption}
\usepackage{booktabs} %

\usepackage{algpseudocode}
\usepackage{algorithm}
\usepackage{mathrsfs}
\usepackage{setspace}

\usepackage{amsmath, amsthm, amssymb, amsfonts, dsfont, mathrsfs, bigints}
\usepackage{mathtools}
\usepackage{enumitem}
\usepackage{tikz}
\usepackage{ISG_style}

\newcommand*\diff{\mathop{}\!\mathrm{d}}


\DeclareMathOperator*{\argmax}{arg\,max}
\DeclareMathOperator*{\argmin}{arg\,min}
\DeclareMathAlphabet\mathbfcal{OMS}{cmsy}{b}{n}

\usepackage{pgfplots}
\pgfplotsset{compat=newest} 
\usepgfplotslibrary{external}
\usepgfplotslibrary{fillbetween}
\usepgfplotslibrary{groupplots}
\usetikzlibrary{patterns}

\usepackage[most]{tcolorbox}

\tcbset{
  aibox/.style={
    width=\linewidth,
    top=7pt,
    bottom=2pt,
    colback=blue!6!white,
    colframe=black,
    colbacktitle=black,
    enhanced,
    center,
    attach boxed title to top left={yshift=-0.1in,xshift=0.15in},
    boxed title style={boxrule=0pt,colframe=white,},
  }
}
\newtcolorbox{AIbox}[2][]{aibox,title=#2,#1}

\usepackage{wrapfig}
\definecolor{lightblue}{rgb}{0.22,0.45,0.70}%
\usepackage{tabularx}

\definecolor{rliableolive}{HTML}{BBCC33}
\definecolor{rliableblue}{HTML}{77AADD}
\definecolor{rliablered}{HTML}{EE8866}

\usepackage{crossreftools}
\usepackage[suppress]{color-edits}

\usepackage{dsfont}
\usepackage{nicefrac}
\newtcolorbox{analysisbox}[1][]{
    enhanced jigsaw,
    colback=white,
    colframe=blue!75!black,
    fonttitle=\bfseries,
    boxsep=5pt,
    left=5pt,
    right=5pt,
    top=5pt,
    bottom=5pt,
    title=#1,
}
\definecolor{editInitialResponse}{RGB}{255, 235, 156} %
\definecolor{editBacktrack}{RGB}{0, 0, 139} %
\definecolor{editRevisedResponse}{RGB}{255, 182, 193} %

\usepackage{pifont}
\usepackage{soul}
\definecolor{highlightmistake}{RGB}{255, 179, 179} 
\definecolor{highlightcorrect}{RGB}{179, 255, 179}

\usepackage[capitalize,noabbrev]{cleveref}

\theoremstyle{plain}
\newtheorem{theorem}{Theorem}[section]

\newtheorem{lemma}[theorem]{Lemma}

\newtheorem{assumption}[theorem]{Assumption}
\theoremstyle{definition}

\theoremstyle{remark}
\newtheorem{remark}[theorem]{Remark}

\usepackage[textsize=tiny]{todonotes}

\tcolorboxenvironment{theorem}{
  colframe=black,
  colback=blue!5!white,
  coltitle=blue,
  boxrule=0.7pt,
  arc=3pt,
  breakable,
}

\tcolorboxenvironment{lemma}{
  colframe=black,
  colback=blue!5!white,
  coltitle=blue,
  boxrule=0.7pt,
  arc=3pt,
  breakable,
}

\usepackage{enumitem}

\usepackage{newfloat}
\usepackage{listings}
\DeclareCaptionStyle{ruled}{labelfont=normalfont,labelsep=colon,strut=off} 
\floatstyle{ruled}
\newfloat{listing}{tb}{lst}{}
\floatname{listing}{Listing}

\usepackage{booktabs}
\usepackage{multirow}
\usepackage{bm}
\usepackage{siunitx}
\let\cite\citep
\title{SharedRep-RLHF: A Shared Representation Approach to RLHF with Diverse Preferences}

\author[$\dagger$]{Arpan Mukherjee$^*$}
\author[$\dagger$]{Marcello Bullo$^*$}
\author[$\dagger$]{Deniz Gündüz}

\affil[$\dagger$]{Imperial College London}

\begin{document}

\maketitle
\def\thefootnote{*}\footnotetext{Equal contribution}\def\thefootnote{\arabic{footnote}}

\allowdisplaybreaks

\vspace{-0.05cm}
\begin{tcolorbox}[
  colback=gray!10,    
  colframe=white,
  boxrule=0.5pt,      
  arc=2pt,            
  left=6pt,right=6pt, 
  top=4pt,bottom=4pt
]
\textbf{Abstract:} Uniform-reward reinforcement learning from human feedback (RLHF), which trains a single reward model to represent the preferences of all annotators, fails to capture the diversity of opinions across sub-populations, inadvertently favoring dominant groups. The state-of-the-art, MaxMin-RLHF, addresses this by learning group-specific reward models, and by optimizing for the group receiving the minimum reward, thereby promoting fairness. However, we identify that a key limitation of MaxMin-RLHF is its poor performance when the minimum-reward group is a minority. To mitigate this drawback, we introduce a novel framework, termed {\em SharedRep-RLHF}. At its core, SharedRep-RLHF learns and leverages {\em shared traits} in annotations among various groups, in contrast to learning separate reward models across groups. We first show that MaxMin-RLHF is provably suboptimal in learning shared traits, and then quantify the sample complexity of SharedRep-RLHF. Experiments across diverse natural language tasks showcase the effectiveness of SharedRep-RLHF compared to MaxMin-RLHF with a gain of up to~$20$\% in win rate.\\

\vspace{.2cm}
\begin{minipage}{\textwidth}
{\footnotesize
\begin{tabular}{rl}
\textbf{Correspondence:} & Arpan Mukherjee (\texttt{a.mukherjee@imperial.ac.uk})\\
& Marcello Bullo (\texttt{bullo.marcello@gmail.com}) \\
\textbf{Code:} & \url{https://github.com/marcellobullo/sharedrep-rlhf}
\end{tabular}
}
\end{minipage}
\end{tcolorbox}

\section{Motivation \& Overview}
The success of large language models (LLMs) is largely attributed to their ability to generate responses which adhere to human values and behavior. The art of steering LLMs to elicit human-aligned responses is known as {\em alignment}, and {\em reinforcement learning from human feedback} (RLHF) serves as the cornerstone for aligning LLMs~\cite{ouyang2022training,bai2022training}. The canonical RLHF pipeline is a three-step process consisting of (i) supervised fine-tuning (SFT), (ii) reward-modeling, and (iii) reinforcement learning (RL) fine-tuning~\cite{wang2023aligning}. In this paper, our focus is on the second and third steps, i.e., reward-modeling and RL fine-tuning. Reward modeling involves fitting a reward model to a {\em preference dataset}, and RL fine-tuning uses policy-gradient methods (e.g., proximal policy optimization (PPO)~\cite{schulman2017proximal} and group relative preference optimization (GRPO)~\cite{shao2024deepseekmath}) to obtain a policy that maximizes a (regularized) average reward objective based on the modeled reward in the second step.  

\paragraph{Uniform versus diverse preference.} In this paper, we focus on {\em offline RLHF}. For a detailed review of offline versus online RLHF, we refer to Appendix~\ref{appendix: canonical RLHF}. Existing investigations on RLHF~\cite{ouyang2022training,bai2022training,stiennon2022learningsummarizehumanfeedback,liu2024aligning,christian2021alignment,xie2024exploratory,pang2024iterative,cen2024value,zhang2024self,casper2023open} mostly focus on a uniform reward model, which is assumed to represent all annotators. This is indicative of a monolithic view of the world, where annotators are assumed to agree on {\em all} aspects, ignoring a vast array of diverse traits intrinsic to annotators from distinct subpopulations. On the contrary, a holistic view of the world captures both subtle and profound nuances in human traits due to individual differences, such as cognitive styles and biases, cultural and societal factors such as language and dialect, ethical and moral values, as well as task-specific differences such as domain expertise and context dependence. In one of the early investigations into the possibility of misalignment of the views of LLMs (to $60$ different US demographic groups)~\cite{santurkar2023whose}, it was concluded that there was {\em significant} misalignment across different demographics, and in many cases, LLMs' views were considerably biased towards certain groups. In order to promote group fairness, \citeauthor{chakraborty2024maxminrlhfalignmentdiversehuman} introduced a framework called {\em MaxMin-RLHF}, which steers the LLM policy to account for the preferences of the worst-case (potentially the minority) group. In this framework, group-specific reward models are learnt from the preference dataset, and the canonical KL-regularized objective for the RL fine-tuning step is replaced by a MaxMin objective, trying to maximize the (regularized) reward for the {\em worst} (lowest reward) group. This framework has been shown to promote fairness with respect to the preferences across diverse groups, and has since been adopted as the de-facto framework for addressing diversity in human preferences from a fairness viewpoint~\cite{ramesh2024group, son2025robust}. A closely related research direction is that of personalized RLHF~\cite{jang2023personalized, chen2024pal,poddar2024personalizing,dong2025personalization}, where the goal is to adapt the LLMs to the preferences of {\em individual} users. Note that we do not aim to solve personalized RLHF, i.e., personalizing to preferences per user. Rather, our focus is on promoting group-fairness in handling distinct preferences across multiple groups. 

\paragraph{Drawbacks of MaxMin-RLHF.}
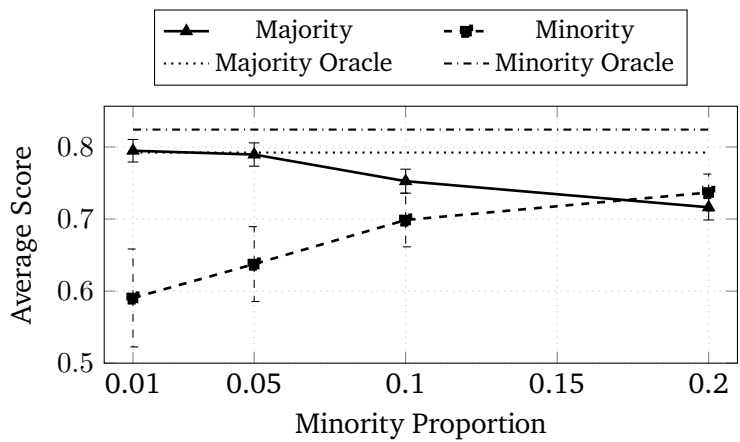
\begin{figure}[t]
    \centering
    \usepgfplotslibrary{fillbetween}
\pgfplotsset{compat=newest, scaled x ticks=false}

\begin{tikzpicture}[scale=1.]
    \begin{axis}[
        xlabel=Minority Proportion,
        ylabel=Average Score,
        xtick={0.01,0.05,0.1,0.15,0.2},
        xticklabels={0.01,0.05,0.1,0.15,0.2},
        ymajorgrids=true,
        xmajorgrids=true,
        grid style=dotted,
        height=5cm,
        width=10cm,
        enlarge x limits=0.05,
        ymin=0.5,
        error bars/y dir=both,
        error bars/y explicit,
        legend style={
            at={(0.5,1.08)},
            anchor=south,
            legend columns=2,
            nodes={scale=0.9, transform shape},
            /tikz/every even column/.append style={column sep=0.6cm}
        },
    ]   
    
    \addplot[
        black,
        line width=1pt,
        mark=triangle*,
        error bars/.cd,
        y dir=both,
        y explicit
    ] table [
        x={Minority Proportion},
        y=mean_score,
        y error=se_margin,
        col sep=comma
    ]{figures/intro_example/data/majority.csv};
    \addlegendentry{Majority}

    \addplot[
        black,
        dashed,
        line width=1pt,
        mark=square*,
        error bars/.cd,
        y dir=both,
        y explicit
    ] table [
        x={Minority Proportion},
        y=mean_score,
        y error=se_margin,
        col sep=comma
    ]{figures/intro_example/data/minority.csv};
    \addlegendentry{Minority}

    \addplot[black, dotted, line width=0.8pt, mark=none] coordinates {(0.01,0.7923) (0.2,0.7923)};
    \addlegendentry{Majority Oracle}

    \addplot[black, dash dot, line width=0.8pt, mark=none] coordinates {(0.01,0.8242) (0.2,0.8242)};
    \addlegendentry{Minority Oracle}

    \end{axis}
    \label{fig:intro_plot}
\end{tikzpicture}
    \caption{Average group scores achieved by MaxMin-RLHF versus varying minority proportions.}
    \label{fig:intro_plot}
\end{figure}
The MaxMin objective, in principle, promotes fairness by steering the policy to cater to the lowest-reward group (reward minority). However, what happens when the reward minority is also a {\em statistical minority}? Does MaxMin-RLHF preserve its group-fairness property when the minority population is sparse? 
In Figure~\ref{fig:intro_plot}, we replicate the ``small scale experiment'' setting in~\cite{chakraborty2024maxminrlhfalignmentdiversehuman} for sentiment analysis using the IMDb dataset~\cite{maas2011learning}, in which the population is split into a {\em majority} and a {\em minority} group. The majority prefers positive reviews, while the minority prefers concise reviews. As the proportion of the minority decreases, Figure~\ref{fig:intro_plot} shows how MaxMin-RLHF alignment generates responses with worse scores for the minority, exhibiting a {\em significant gap} with the oracle policy, which assumes to know the ground truth reward model. MaxMin-RLHF requires a minimal level of minority share ($\approx20 \%$) before it can reach some form of parity with the majority. These effects demonstrate that MaxMin-RLHF is sample-inefficient, suboptimal for both subgroups, and poorly suited to skewed data settings.

\paragraph{Contributions.} We address these limitations of the MaxMin-RLHF framework in settings where minority annotators are underrepresented. Our key insight is that, despite group-specific nuances, preference data across groups often reflect common underlying human values. We propose to learn these {\em shared values} from the entire dataset, while capturing group-specific nuances using annotations from each group. Our contributions are threefold.

\textit{Framework:} We introduce a novel framework, {\em shared representation RLHF} (SharedRep-RLHF), that addresses the estimation inaccuracies of MaxMin-RLHF by learning and leveraging shared traits among groups. The goal, in this premise, is to solve the MaxMin objective to produce a policy that promotes fairness across groups in its generated responses. SharedRep-RLHF is a generalization of the MaxMin-RLHF framework, which is recovered in the special case when groups do not share any common trait. Furthermore, we propose an algorithm called {\em SharedRep-RLHF} that provably optimizes the MaxMin objective proposed in~\cite{chakraborty2024maxminrlhfalignmentdiversehuman}.

\textit{Analytical contributions:} We statistically characterize various algorithmic facets of SharedRep-RLHF. We make three analytical contributions. First, we derive confidence sets for maximum likelihood estimates (MLEs) in the SharedRep-RLHF framework, which scale inversely with the size of the {\em entire dataset} (Lemma~\ref{lemma: est_con_sharedrep}), as opposed to the size of group-specific datasets, which has been reported in~\cite{chakraborty2024maxminrlhfalignmentdiversehuman}. Second, building on this result, we show that SharedRep-RLHF offers provably better estimation fidelity compared to MaxMin-RLHF (Theorem~\ref{theorem: performance comparison}). Third, we characterize the sample complexity of SharedRep-RLHF in the probabably approximately correct (PAC) framework in Theorem~\ref{theorem: sample complexity}. We find that the additional number of samples required to enforce the MaxMin objective over the canonical regularized reward objective~\cite{pmlr-v235-xiong24a} is of the order~$O\Big(\frac{1}{\Delta_{\min}^4}\Big)$, where $\Delta_{\min}$ is the minimum gap between conditional entropies of the induced Gibbs distributions of the groups, and has been formally defined in~\eqref{eq:Delta}. In the process, we also show that the group that has the {\em minimum reward} is equivalent to the group exhibiting {\em maximal entropy} in its induced Gibbs distribution (Lemma~\ref{lemma:equivalence_of_u}). Note that the framework in~\cite{chakraborty2024maxminrlhfalignmentdiversehuman} does not come with statistical guarantees; hence, we provide the first result on the sample complexity of MaxMin-RLHF.

\textit{Experiments:} We empirically validate SharedRep-RLHF on three diverse tasks -- {\em controlled sentiment analysis}, {\em mathematical reasoning}, and {\em single turn dialogue} -- under varying degrees of minority underrepresentation. SharedRep-RLHF consistently outperforms MaxMin-RLHF in both mean minority score and win rate, especially in low minority proportion regimes, demonstrating its effectiveness and robustness in aligning the underlying LLM with underrepresented group preferences leveraging shared values across group prefernces.

\section{Preliminaries}
In this section, we provide a brief background on the MaxMin-RLHF pipeline~\cite{chakraborty2024maxminrlhfalignmentdiversehuman} and subsequently introduce our proposed SharedRep-RLHF framework, a generalization of MaxMin-RLHF that is cognizant of shared preferences among the groups. 

\paragraph{MaxMin-RLHF.} The premise in RLHF is that the learner has access to a preference dataset denoted by $\mcD \triangleq \{ (\bx_i,\by_i,\by^\prime_i,z_i)\}_{i=1}^N$. Here, for each prompt $\bx_i$, $i\in[N]$, $\by_i,\by_i^\prime\sim\pi_{\rm ref}(\cdot\med\bx_i)$ denote two responses generated by the reference model conditioned on prompt $\bx_i$, and $z_i\in\{0,1\}$ denotes the annotators' preference, where $z_i=1$ if the annotator prefers $\by_i$ as a response to prompt $\bx_i$, and $z_i=0$ otherwise. We denote the space of prompts by $\mcX$ and the space of responses by $\mcY$. In order to capture the diversity in human preferences,~\citeauthor{chakraborty2024maxminrlhfalignmentdiversehuman} assume that the annotators can be clustered into $U$ human subpopulations. Let $\mcH$ denote the set of all annotators. Accordingly, we assume that $\mcH\;=\;\displaystyle\cup_{u=1}^U \mcH_u$ such that $\mcH_u\displaystyle\cap\mcH_v\;=\;\emptyset$ for all $u\neq v$. Based on the partitioning of the annotators,~\citeauthor{chakraborty2024maxminrlhfalignmentdiversehuman} assume that each subpopulation $u\in[U]$ has an intrinsic parameter $\btheta_u^\star$ that models its preferences using the Bradley-Terry (BT) model~\cite{bradley1952rank}, i.e., 
\begin{align}
\label{eq:group_BTL}
    \P_u\Big (z = 1 \med \bx,\by,\by^\prime\Big )\;\triangleq\; \sigma \Big( r_{\btheta^\star_u}(\bx,\by) - r_{\btheta^\star_u}(\bx,\by^\prime)\Big)\ ,
\end{align}
where $\sigma$ denotes the sigmoid function, $\P_u$ denotes the preference measure corresponding to the subpopulation $u\in[U]$ and $r_{\btheta^\star_u}(\cdot , \cdot)$ denotes a parametric representation of its intrinsic reward function $r_u^\star(\cdot,\cdot)$ guiding the preferences. We further assume that the partitioning of the annotators is {\em known}, i.e., for each annotator $i\in\mcH$, we know the subpopulation to which they belong, i.e., $i\in\mcH_u$\footnote{Alternatively, we can always invoke~\cite[Algorithm $1$]{chakraborty2024maxminrlhfalignmentdiversehuman} to learn an appropriate clustering.}, $u\in[U]$. In this setting, distinctly to uniform preference RLHF, we can form estimates of group-specific parameters $\widehat\btheta_u$ for every $u\in[U]$ by performing reward-modeling using group-specific preference data. Once we form these estimates,~\citeauthor{chakraborty2024maxminrlhfalignmentdiversehuman} propose to solve the MaxMin reward objective:
\small{
\begin{align}
    \pi_{\rm MaxMin}\;\in\; &\argmax_\pi\;\bigg\{ \min\limits_{u\in[U]} \E_{\bx\sim\rho, \by\sim\pi(\cdot\med \bx)} \Big[ r_{\widehat\btheta_u}(\bx,\by) -\beta D_{\sf KL}\big(\pi(\cdot\med \bx)\|\pi_{\rm ref}(\cdot\med\bx) \big)\Big]\bigg\}\ ,
\end{align}}
inspired by the egalitarian principle in social choice theory~\cite{Sen2017Collective}. The objective corresponds to maximizing the regularized value function corresponding to the {\em minimum-reward} subpopulation.


\paragraph{SharedRep-RLHF.}~\citeauthor{chakraborty2024maxminrlhfalignmentdiversehuman} assume that the subpopulation-specific preferences are completely disparate, and the groups do not share any commonalities in their preferences. However, this is not true in general. There are indeed prompts and answers which share a common preference across all subpopulations. Hence, there exist common traits across the whole population $\mcH$ that can be estimated from the preference data across all subpopulations. We propose to view the problem from the lens of representation learning in bandits~\cite{yangimpact}. Specifically, we assume that there is a universal feature extractor matrix $\bB^\star\in\R^{d\times K}$ across subpopulations, whose columns may be interpreted as representing various human traits (usually $K\ll d$). For each subpopulation $u\in[U]$, the intrinsic parameters may be expressed as $\btheta^\star_u\;\triangleq\; \bB^\star\bw^\star_u$,
where $\bw^\star_u\in\Delta^{K-1}$ is a mixing coefficient in the probability simplex of order $K$. We call this framework {\em shared representation RLHF} (SharedRep-RLHF) in light of the common feature extractor $\bB^\star$. In our formulation, for those traits that are shared among the entire population, the mixing coefficients (or weights) would be similar for all subpopulations. Subpopulation-specific traits, on the other hand, would exhibit disparity in their weights learned across various subpopulations. Note that this framework provides greater flexibility compared to~\cite{chakraborty2024maxminrlhfalignmentdiversehuman}. To elaborate, when we have fewer preference data for a minority subpopulation, attempting to estimate the corresponding intrinsic parameter from limited data points might result in extremely inaccurate or unreliable reward estimates. In the proposed SharedRep-RLHF framework, data from the entire population can be used to estimate the feature extractor. Hence, we hope to achieve better estimation accuracy leveraging the feature extractor compared to the framework in~\cite{chakraborty2024maxminrlhfalignmentdiversehuman}. In order to statistically analyze the performance of the proposed algorithm, we make the following assumptions which are commonly adopted for the analysis of RLHF algorithms.

\begin{assumption}[Linear reward]
\label{assumption:linear reward}
    We assume that the reward function is linearly parameterized, i.e., for each subpopulation $u\in[U]$, we have $r_{\bB\bw_u}(\bx,\by)\triangleq \big\langle \phi(\bx,\by), \bB\bw_u\big\rangle$ for any feature extractor $\bB\in\mcB$ and $\bw_u\in\Delta^{K-1}$, such that $\mcB\triangleq\big\{\bB \in\R^{d\times K} : \|\bB(:,k)\|_2\leq B_{\max}\ ,\;\forall\;k\in[K]\big\}$, $\bB(:,k)$ denotes the $k^{\rm th}$ column of matrix $\bB$, and $\phi(\bx,\by)$ denotes a known embedding of the concatenated (prompt, response) pair $(\bx,\by)$. Furthermore, we assume that $\|\phi(\cdot,\cdot)\|\leq L_{\max}$.
\end{assumption}
Assumption~\ref{assumption:linear reward} helps in terms of a clear presentation of our results, and has been extensively used for deriving statistical guarantees in the RLHF literature~\cite{kong2022provably,saha2023dueling,zhu2023principled,pmlr-v235-xiong24a,foster2025good}. In experiments, $\phi$ is generally chosen as the final layer of the frozen SFT model, after removing the logits and the softmax layers. In contrast to the existing norm of adding a single neuron after the last layer of the frozen SFT backbone with fully connected and trainable weights, the SharedRep-RLHF framework adds a shared linear layer $\bB$, which captures the shared preferences between subpopulations, followed by a fully connected neuron for each subpopulation. We refer to Figure~\ref{fig:architecture} in Appendix~\ref{appendix: experiments} for a detailed overview of the architecture in comaprison with MaxMin-RLHF.

\begin{assumption}[Reward gap]
\label{assumption:reward gap}
    For any subpopulation $u\in[U]$, and for any pair of parameters in the hypothesis classes $\mcB$ and $\Delta^{K-1}$, there exists a prompt for which the minimum reward difference between the chosen and rejected responses is bounded away from $0$. Specifically, we assume that $\xi_u>0$ for all $u\in[U]$, where we have defined
    \small{\begin{align}
        \xi_u\triangleq \inf\limits_{\bB\in\mcB,\bw_u\in\Delta^{K-1}}\;\max\limits_{\bx\in\mcX}\;\min\limits_{\by\neq\by^\prime}\;\Big |\big\langle \phi(\bx,\by) - \phi(\bx,\by^\prime),\bB\bw_u\big\rangle \Big |\ .
    \end{align} }
\end{assumption}
Assumption~\ref{assumption:reward gap} implies that for {\em every} model in the parameter space, there always exist a prompt which would yield a minimum reward difference that is bounded away from $0$. This is a very natural assumption, as otherwise we would have the possibility of models under which it is {\em impossible to differentiate} between distinct pairs of responses. In the reward-modeling stage, we form estimates $\widehat\bB$ for the feature extractor $\bB^\star$, and $\widehat\bW\triangleq[\widehat\bw_1,\cdots,\widehat\bw_U]$ corresponding to the weights $\bW^\star\triangleq[\bw_1^\star,\cdots,\bw_U^\star]$. Finally, we assume that there are no degenerate states. 
\begin{assumption}[Non-degeneracy]
\label{assumption:non_degeneracy}
    The distribution over prompts, denoted by $\rho$, satisfies $\rho_{\min}>0$, where we have defined $\rho_{\min}\triangleq\min_{\bx\in\mcX}\rho(\bx)$.
\end{assumption}
Next, similar to MaxMin-RLHF, our goal is to design a policy that maximizes the KL-regularized value function comprising the estimated reward models. Specifically, the learner's goal is to design a policy that maximizes the KL-regularized MaxMin value function, i.e., 
\small{
\begin{align}
\label{eq:KL_opt_pol_def}
    \pi^\star\in\argmax_\pi\, J_{\rm MaxMin}(\pi)\triangleq \min\limits_{u\in[U]} \bigg\{\E_{\bx\sim\rho, \by\sim\pi(\cdot\med \bx)} \Big[ r_{\btheta^\star_u}(\bx,\by) -\beta D_{\sf KL}\big(\pi(\cdot\med \bx)\|\pi_{\rm ref}(\cdot\med\bx) \big)\Big]\bigg\}.
\end{align}}

\section{Estimator \& Performance Comparison}
In this section, we propose an estimator for the reward-modeling stage, which is subsequently used to optimize the MaxMin objective. Furthermore, we asses the performance of the proposed estimator, comparing it to the MaxMin-RLHF framework proposed in~\cite{chakraborty2024maxminrlhfalignmentdiversehuman}, and showcasing a provable gain over the MaxMin framework. All proofs are deferred to the appendix for brevity.

\paragraph{Definitions.} Let us first introduce a few notations. For specifying the estimator, for any data point $i\in[N]$, let us define the binary cross-entropy loss
\small{
\begin{align}
    &\ell_i(\bB, \bw_u)\;\triangleq\; z_i\log\left( \sigma\Big(r_{\bB\bw_u}(\bx_i,\by_i) - r_{\bB\bw_u}(\bx_i,\by^\prime_i)\Big)\right) +(1-z_i) \log\left( 1 - \sigma\Big(r_{\bB\bw_u}(\bx_i,\by_i) - r_{\bB\bw_u}(\bx_i,\by^\prime_i)\Big)\right)
    \ .
\end{align}}
The MLE is obtained by minimizing the aggregate loss, i.e., 
\begin{align}
\label{eq: sharedrep_estimator}
    \widehat\bB\;,\;\widehat\bW\;\in\; \argmin_{\bB\in\mcB, \bw_u\in\Delta^{K-1},\;\forall u\in[U]}\; \sum\limits_{i\in[N]} \ell_i(\bB, \bw_u)\ .
\end{align}
In order to capture the estimation fidelity based on the offline dataset $\mcD$, following prior art~\cite{zhu2023principled,das2024active}, we adopt the {\em unregularized} value function as the performance metric. Specifically, for any policy $\pi$, and for any subpopulation $u\in[U]$, let us define the unregularized value function under the ground truth as
\begin{align}
\label{eq:GT_unregularized_value_function}
    J_u(\pi)\;\triangleq\; \E_{\bx\sim\rho,\by\sim\pi(\cdot\med\bx)}\Big[ r_{\btheta^\star_u}(\bx,\by)\Big]\ .
\end{align}
Furthermore, let $\pi^\star_u \in \argmax_{\pi} J_u(\pi)$ denote the optimal policy maximizing the unregularized value function for subpopulation $u\in[U]$. Let ${\mathfrak{A}}$ denote an algorithm used for forming estimates of the model parameters. The fidelity of estimation is captured through the value function gap due to the induced policy $\widehat\pi_{\mathfrak{A}}$, i.e., for any group $u\in[U]$,
\begin{align}
    {\rm SubOpt}_u(\widehat\pi_{\mathfrak{A}})\;\triangleq\; J_u(\pi^\star_u) - J_u(\widehat\pi_{\mathfrak{A}})\ .
\end{align}

\paragraph{Pessimism in offline RLHF.} It is well-established in the offline RLHF literature~\cite{zhu2023principled,pmlr-v235-xiong24a} that using MLEs (i.e., $\mathfrak{A} = {\rm MLE}$) may result in a potentially unbounded suboptimality gap. 
A mechanism to handle this issue is the principle of {\em pessimism}. Specifically, instead of relying entirely on the MLE, we form pessimistic estimates based on appropriately designed confidence sets around the MLEs. To elaborate on this, we first introduce confidence sequences for MLEs under the MaxMin-RLHF framework as well as the SharedRep-RLHF framework. Let us split the dataset $\mcD$ based on annotations according to subpopulations $u\in[U]$, and let $N_u = |\mcH_u|$ denote the number of data points annotated by annotators from subpopulation $u$. The confidence sequence constructed based on the MaxMin-RLHF framework follows from~\cite[Lemma 3.1]{zhu2023principled}, and is stated below. Note that we use ${\mathfrak{A}} = {\rm MM}$ and ${\mathfrak{A}} = {\rm SR}$ to denote the MaxMin-RLHF and the SharedRep-RLHF frameworks.
\begin{lemma}[\citet{zhu2023principled}]
\label{lemma: est_con_maxmin}
    Recall that $\widehat\btheta_u$ denotes the MLE formed under the MaxMin-RLHF framework from the subpopulation-specific data for each $u\in[U]$. Under Assumption~\ref{assumption:linear reward}, for any $\lambda\in\R_+$ and $\delta\in(0,1)$, with probability at least $1-\delta$ we have
    \begin{align}
        \big\|\widehat\btheta_u - \btheta_u^\star \big\|_{\Sigma_u + \lambda \mathbb{I}}\;\leq\; C_{\rm MM}\sqrt{\frac{1}{N_u}C_{\delta}+\lambda B_{\max}^2}\ ,
    \end{align}
    where $\Sigma_u\triangleq \frac{1}{N_u}\sum_{i\in\mcH_u} \big( \phi(\bx_i,\by_i) - \phi(\bx_i,\by_i^\prime)\big)\big( \phi(\bx_i,\by_i) - \phi(\bx_i ,\by_i^\prime)\big)^\top$, $C_{\delta}\triangleq \frac{d+\log(1/\delta)}{\gamma^2}$, $\gamma \triangleq 1/\big(2 + \exp(-L_{\max}B_{\max}) + \exp(L_{\max}B_{\max})\big)$, and $C_{\rm MM}\in\R_+$ is a universal constant. 
\end{lemma}
\begin{remark}
    We observe from Lemma~\ref{lemma: est_con_maxmin} that the accuracy of subpopulation-specific reward estimates rely heavily on the size of the subpopulation $N_u$. Specifically, if $N_u$ is small, the estimation error $\|\widehat\btheta_u - \btheta_u^\star \|_{\Sigma_u + \lambda \mathbb{I}}$ is only controlled in the few directions covered by $\Sigma_u$, and the confidence bound is relatively large. This matches our experimental results in Figure~\ref{fig:intro_plot}: the smaller the subpopulation size, the worse the estimation accuracy, and consequently, the worse the performance of the MaxMin-RLHF framework.
\end{remark}
Unlike the MaxMin-RLHF framework, we propose to learn traits shared among subpopulations, and as a result, improve the estimation accuracy over MaxMin-RLHF. The SharedRep-RLHF framework learns a shared feature extractor $\bB$ from the entire dataset $\mcD$. This aids in the estimation of $\btheta_u^\star$ for every $u\in[U]$, since we are no longer restricted by directions which are exclusively covered by $\Sigma_u$. More specifically, we have the following concentration on the MLEs in the SharedRep-RLHF framework. 
\begin{lemma}[Estimator concentration]
\label{lemma: est_con_sharedrep}
    Under Assumption~\ref{assumption:linear reward}, for any $\lambda\in\R_+$ and $\delta\in(0,1)$, with probability at least $1-\delta$, for every subpopulation $u\in[U]$ we have
    \begin{align}
        \big\| \widehat\bB\widehat\bw_u - \bB^\star\bw_u^\star\big\|_{\Sigma + \lambda\mathbb{I}}\;\leq\; C_{\rm SR}\sqrt{\frac{1}{N} C_{\delta}+\lambda B_{\max}^2}\ ,
    \end{align}
    where $\Sigma\triangleq \frac{1}{N}\sum_{i\in[N]} \big( \phi(\bx_i,\by_i) - \phi(\bx_i ,\by_i^\prime)\big)\big( \phi(\bx_i,\by_i) - \phi(\bx_i ,\by_i^\prime)\big)^\top$, and $C_{\rm SR}\in\R_+$ is a universal constant.
\end{lemma}
\paragraph{Pessimistic value functions.} Based on Lemmas~\ref{lemma: est_con_maxmin} and~\ref{lemma: est_con_sharedrep}, we now define confidence sets, which indicate the estimation fidelity in the MaxMin and SharedRep frameworks, respectively, as follows. For every $u\in[U]$,
\begin{align}
    \Theta_{\rm MM}(\widehat\btheta_u)\;&\triangleq\;\Big\{\btheta\in\Theta : \|\widehat\btheta_u - \btheta \|_{\Sigma_u + \lambda \mathbb{I}}\; \leq\; C_{\rm MM}\sqrt{\frac{1}{N_u}C_{\delta}+\lambda B_{\max}^2} \Big\}\ ,\\
    \Theta_{\rm SR}(\widehat\bB,\widehat\bw_u)\;&\triangleq\;\Big\{ (\bB,\bw)\in\mcB\times\Delta^{K-1} : \| \widehat\bB\widehat\bw_u - \bB\bw\|_{\Sigma + \lambda\mathbb{I}}\; \leq\; C_{\rm SR}\sqrt{\frac{1}{N}C_{\delta}+\lambda B_{\max}^2}\Big\}\ .
\end{align}
Furthermore, we define {\em pessimistic value functions} based on the above confidence sequences in each framework for any policy $\pi$. The pessimism comes from the parameter estimate, which is chosen as the one that yields the {\em minimal value function} within the confidence sets. Formally, in the MaxMin-RLHF framework, for a policy $\pi$ and for any subpopulation $u\in[U]$, we define the estimated pessimistic value function and its optimal policy as
\begin{align}
    \widehat J_u^{\rm MM}(\pi)\;\triangleq\; \min_{\btheta\in\Theta_{\rm MM}(\widehat\btheta_u)}\; \E_{\bx\sim\rho,\by\sim\pi(\cdot\med\bx)}\Big[ \big\langle \phi(\bx,\by),\btheta\big\rangle\Big]\ ,\quad \textrm{and} \quad \widehat\pi_u^{\rm MM} \in\; \argmax\limits_{\pi}\; \widehat J_u(\pi)\ .
\end{align}
Similarly, in the SharedRep-RLHF framework, for any $u\in[U]$, we define the estimated pessimistic value function and its optimal policy as
\begin{align}
    \widehat J_{u}^{{\rm SR}}(\pi)\;\triangleq\; \min\limits_{(\bB,\bw)\in\Theta_{\rm SR} (\widehat\bB,\widehat\bw_u)}\; \E_{\bx\sim\rho,\by\sim\pi(\cdot\med\bx)}\Big[ \big\langle \phi(\bx,\by),\bB\bw\big\rangle\Big]\ ,\quad \textrm{and} \quad \widehat\pi_u^{\rm SR}\in\; \argmax\limits_{\pi}\; \widehat J_u^{\rm SR}(\pi)\ .
\end{align}
Note that 
prior work~\cite{zhu2023principled} proposed defining value functions shifted by a fixed bias vector $\bnu$. Our results remain valid under this formulation; however, for clarity of presentation, we omit the bias vector in our definitions.

\paragraph{Performance comparison.} Having defined the estimated value functions $\widehat J_u^{\rm MM}$ and $\widehat J_u^{\rm SR}$, we now demonstrate that the estimation framework of SharedRep-RLHF offers provably improved performance over MaxMin-RLHF. This comparison is made using the pessimistic value function as the performance metric. Specifically, we have the following theorem.
\begin{theorem}[Performance comparison]
\label{theorem: performance comparison}
    Under Assumptions~\ref{assumption:linear reward},\ref{assumption:reward gap}, and~\ref{assumption:non_degeneracy}, if $\widehat\pi_u^{\rm MM} \neq \widehat\pi_u^{\rm SR}$, for any $u\in[U]$ and for any $\delta\in(0,1)$, with probability at least $1-\delta$ we have
    \begin{align}
        &{\rm SubOpt}_u(\widehat\pi_u^{\rm MM}) - {\rm SubOpt}_u(\widehat\pi_u^{\rm SR})\; \geq\; \rho_{\min}\xi_u - 2\eta^{\rm SR}(N,\lambda,\delta)\cdot\E_{\bx\sim\rho}\left [ \kappa^{\rm MM}(\Sigma,\lambda,\bx)\right]\ , 
    \end{align}
    where we have defined 
    $$\eta^{\rm SR}(N,\lambda,\delta)\triangleq C_{\rm SR}\sqrt{\frac{1}{ N} C_\delta + \lambda B_{\max}^2}\ ,$$ and 
    $$\kappa^{\rm MM}(\Sigma,\lambda,\bx)\triangleq \big\| \E_{\by\sim\widehat\pi_u^{\rm MM}(\cdot\med\bx)}\big[\phi(\bx,\by)\big]\big\|_{(\Sigma + \lambda\mathbb{I})^{-1}}\ .$$
\end{theorem}
Theorem~\ref{theorem: performance comparison} shows that when $\lambda = \frac{1}{N}$, as the dataset size grows, the performance gain of SharedRep-RLHF estimation becomes evident. Asymptotically, the performance gap is quantified through a constant term $\xi_u$ for any subpopulation $u\in[U]$, which captures the (prompt-wise) maximal minimum reward gap between the chosen and rejected responses that exists in the dataset. Additionally, the term $\left\| \E_{\by\sim\widehat\pi_u^{\rm MM}(\cdot,|,\bx)}\big[\phi(\bx,\by)\big] \right\|_{(\Sigma + \lambda\mathbb{I})^{-1}}$ can be upper-bounded by $L_{\max} \sqrt{\lambda_{\min}(\Sigma + \lambda\mathbb{I})}$, where $\lambda_{\min}(\bA)$ denotes the smallest eigenvalue of matrix $\bA$. Consequently, this multiplicative factor is, in the worst-case, $O(1)$.

\section{Sample Complexity of SharedRep-RLHF}

In this Section, we provide an algorithm for optimizing the {\em KL-regularized reward objective} stated in~\eqref{eq:KL_opt_pol_def} in the SharedRep-RLHF framework. In canonical RLHF, GRPO and PPO are used to find a near-optimal solution to~\eqref{eq:KL_opt_pol_def}, assuming that $U=1$. We extend this by proposing an algorithm tailored to optimize the KL-regularized value function in~\eqref{eq:KL_opt_pol_def} across multiple subpopulations. After the reward learning phase, the algorithm applies GRPO (or PPO) using the value function associated with the subpopulation exhibiting the lowest average reward, similarly to MaxMin-RLHF. Furthermore, while \citeauthor{chakraborty2024maxminrlhfalignmentdiversehuman} does not provide performance guarantees for MaxMin-RLHF, we provide a sample complexity analysis of the proposed method, establishing conditions under which it achieves an $\varepsilon$-accurate approximation of the optimal value function. 

\begin{figure}[h]
    \begin{minipage}{\textwidth}
    \begin{algorithm}[H]
        \small
        \caption{SharedRep-RLHF Algorithm}
        \label{algorithm:SharedRep}
        \begin{algorithmic}[1]
            \State \textbf{Input:} Dataset $\mcD$, regularization $\lambda$, $B_{\max}$, $K$
            \State Compute MLEs $\widehat\bB$ and $\widehat\bW$ based on~\eqref{eq: sharedrep_estimator}
            \State $\Gamma(\Sigma,\lambda,\pi)\triangleq \|\E_{\bx\sim\rho,\by\sim\pi(\cdot\med\bx)}
            [\phi(\bx,\by)]\|_{(\Sigma+\lambda\mathbb{I})^{-1}}$
            \State $\eta^{\rm SR}(N,\lambda,\delta)\triangleq 
            C_{\rm SR}\sqrt{
            \frac{d+\log(1/\delta)}{\gamma^2 N} + \lambda B_{\max}^2}$
            \State \textbf{Output:}
            \[
            \begin{aligned}
            \widetilde\pi^{\rm SR} \in \argmax_{\pi} &\min_{u \in [U]} \Big\{ 
            \E_{\substack{\bx \sim \rho \\ \by \sim \pi(\cdot \mid \bx)}}
            \left[r_{\widehat\bB \widehat\bw_u}(\bx,\by)\right] - \eta^{\rm SR}(N,\lambda,\delta)\,\Gamma(\Sigma,\lambda,\pi) - \beta D_{\sf KL}\left(\pi(\cdot \mid \bx) \| \pi_{\rm ref}(\cdot \mid \bx)\right)
            \Big\}
            \end{aligned}
            \]
        \end{algorithmic}
    \end{algorithm} 
    \end{minipage}
\end{figure}

\paragraph{SharedRep-RLHF Algorithm.} We propose the SharedRep-RLHF algorithm which is based on the principles of MaxMin-RLHF. First, we form MLEs of the shared and subpopulation-specific parameters $\widehat\bB$, and $\widehat\bW$, respectively. These estimates guide in forming appropriate confidence sets based on Lemma~\ref{lemma: est_con_sharedrep}. After reward-modeling, we leverage the MaxMin objective to align the SFT model based on pessimistic reward estimates from the previous step. In order to implement pessimism, we adopt the technic proposed in~\cite{pmlr-v235-xiong24a}, which proposes to subtract an uncertainty term $\Gamma(\Sigma,\lambda,\pi)$ scaled by the confidence width $\eta^{\rm SR}(N,\lambda,\delta)$ from the average reward under a policy $\pi$. Finally, we adopt the MaxMin objective to find a policy that maximizes the worst-case pessimistic KL-regularized reward. The entire pseudocode is presented in Algorithm~\ref{algorithm:SharedRep}.  

    

\paragraph{Performance Guarantees.} The SharedRep-RLHF algorithm involves finding a policy $\pi$ that maximizes the estimated average reward corresponding to the {\em worst-case subpopulation}, i.e.,
\begin{align}
\label{eq:u_hat_reward}
    \widehat u\;\in\;\argmin\limits_{u\in[U]} \;\E_{\bx\sim\rho,\by\sim\widetilde\pi^{\rm SR}(\cdot\med\bx)}\Big[ r_{\widehat\bB\widehat\bw_u}(\bx,\by)\Big]\ .
\end{align}
In order to provide a sample complexity analysis of the SharedRep-RLHF algorithm, we first make an observation that the subpopulation corresponding to the worst-case reward, as stated in~\eqref{eq:u_hat_reward}, can equivalently be expressed as the subpopulation that maximizes the entropy of the intrinsic conditional Gibbs distribution. In order to formalize this, we introduce some notations. For any set of reward functions $\{r_u : u\in[U]\}$, let us define the induced MaxMin policy as
\begin{align}
\label{eq:pi_r}
    \pi_r\;&\in\;\argmax_\pi\;\min_{u\in[U]}\;\bigg\{\E_{\bx\sim\rho,\by\sim\pi(\cdot\med\bx)} \Big[r_u(\bx,\by)\Big]- \beta D_{\sf KL}\big(\pi(\cdot\med\bx)\|\pi_{\rm ref}(\cdot\med\bx) \big)\bigg\}\ .
\end{align}
Furthermore, let $u_r$ denote the corresponding worst-case subpopulation, i.e.,
\begin{align}
\label{eq:u_r}
    u_r\;\in\;\argmin\limits_{u\in[U]} \;\E_{\bx\sim\rho,\by\sim\pi_r(\cdot\med\bx)}\Big[ r_u(\bx,\by)\Big]\ .
\end{align}
Let us define the Gibbs distributions induced by a reward function $r$ given a prompt $\bx$ as
\begin{align}
\label{eq:Gibbs_general}
    \nu_r(\by\med\bx)\;\triangleq\; \frac{\pi_{\rm ref}(\by\med\bx)\exp\left( \frac{1}{\beta}r(\bx,\by)\right)}{\sum_{\by} \pi_{\rm ref}(\by\med\bx)\exp\left( \frac{1}{\beta}r(\bx,\by)\right)}\ .
\end{align}
Furthermore, let $H(\nu_r(\cdot\med\bX))$ denote the conditional entropy of the Gibbs distribution corresponding to the subpopulation $u\in[U]$, i.e., $H(\nu_r(\cdot\med\bX))\triangleq \E_{\bx\sim\rho}[H(\nu_r(\cdot\med\bx))]$. In the following lemma, we establish the equivalence between selecting the subpopulation that exhibits the minimal average reward, and selecting the one that maximizes the (conditional) entropy of its Gibbs distribution.
\begin{lemma}[Worst-case subpopulation]
\label{lemma:equivalence_of_u}
    The worst-case subpopulation $u_r$ obtained from solving the MaxMin objective for a set of reward functions $\{r_u : u\in[U]\}$, given in~\eqref{eq:u_r}, is equivalent to solving for the subpopulation exhibiting the maximum conditional entropy of the Gibbs distribution, i.e., $u_r\;\in\;\argmax\limits_{u\in[U]}\; H(\nu_{r_u}(\cdot\med\bX) )\ .$
\end{lemma}
We leverage Lemma~\ref{lemma:equivalence_of_u} to derive an upper bound on the sample complexity of the SharedRep-RLHF algorithm. Specifically, we will quantify a {\em sufficient} number of samples $N$ that ensures an $(\varepsilon,\delta)$-PAC guarantee on the estimated regularized MaxMin value function. In order to specify this, we introduce a few notations. For any $u\in[U]$, let $\nu^\star_u(\cdot\med\bx)$ denote the Gibbs distribution induced by the true reward model $r_{\btheta_u^\star}$ as specified in~\eqref{eq:Gibbs_general}. Furthermore, let $u^\star$ denote the worst-case subpopulation obtained by plugging the ground truth rewards $\{r_{\btheta^\star_u}: u\in[U]\}$ in~\eqref{eq:u_r}. Accordingly, we denote the suboptimality gaps
\begin{align}
\label{eq:Delta}
    \Delta_u\;\triangleq\; \Big |H\big( \nu^\star_{u^\star}(\cdot\med\bX)\big) - H\big( \nu^\star_u(\cdot\med\bX)\big)\Big |\;\;\forall\;u\neq u^\star\ , \quad \textrm{and} \quad \Delta_{\min}\;\triangleq\;\min\limits_{u\in[U]\setminus \{u^\star\}}\;\Delta_u\ .
\end{align}
In~\eqref{eq:Delta}, $\Delta_u$ captures the gap in the conditional entropy of the Gibbs distributions corresponding to the worst-case subpopulation $u^\star$ and any other subpopulation $u\neq u^\star$. The minimum gap $\Delta_{\min}$ captures the hardness of identifying the worst-case subpopulation under the ground truth from the other subpopulations. Furthermore, for each $u\in[U]$ we define dataset-specific parameters
\begin{align}
    \psi_u(\Sigma, \beta, \delta)\;&\triangleq\; \frac{1}{\beta^2}(C_\delta + B_{\max}^2) \times\left(\max_{\bx\in\mcX}\;\E_{\by\sim\nu^\star_u(\cdot\med\bx)}\Big[ \|\phi(\bx,\by) \|_{\big( \Sigma + \lambda\mathbb{I}\big)^{-1}}\Big]\right)^2\ .
\end{align}
Based on these definitions, we provide the following guarantee on the sample complexity of the SharedRep-RLHF algorithm.
\begin{theorem}[Sample Complexity]
    \label{theorem: sample complexity}
    Let us set $\lambda = \frac{1}{N}$. Under Assumption~\ref{assumption:linear reward}, 
    \begin{align}
    \label{eq:theorem_sample_complexity}
        &N^{\rm SR}\;\triangleq\;\max\bigg\{ N_{\rm MaxMin}, \; O\left( \frac{C_\delta}{\varepsilon^2}\big\|\E_{\bx\sim\rho,\by\sim\pi^\star(\cdot\med\bx)}\big[ \phi(\bx,\by)\big] \big\|^2_{\big(\Sigma + \lambda\mathbb{I}\big)^{-1}}\right)\bigg\}
    \end{align}
    samples are sufficient to ensure that $\P\big( J_{\rm MaxMin}(\pi^\star) - J_{\rm MaxMin}(\widetilde\pi^{\rm SR})\leq\varepsilon\big)\;>\;1-\delta$,
    where $N_{\rm MaxMin}$ is defined as follows.
    \begin{enumerate}
        \item \textbf{(Large-gap regime.)} If $\Delta_{\min} > \frac{2}{\e}\big(\log|\mcY| +2\big)$, we define 
        \begin{align}
            N_{\rm MaxMin}\;\triangleq\;\max_{u\in[U]}\bigg\{ O\bigg( \psi_u(\Sigma, \beta, \delta)\Big(\frac{\log|\mcY|+2}{\Delta_{\min}}\Big)^4\bigg)\bigg\}\ .
        \end{align}
        \item \textbf{(Small-gap regime.)} If $\Delta_{\min} \leq \frac{2}{\e}\big(\log|\mcY| +2\big)$, we define
        \begin{align}
            N_{\rm MaxMin}\;&\triangleq\;\max_{u\in[U]}\bigg\{O\bigg( \psi_u(\Sigma, \beta, \delta)\exp\bigg( -4W_{-1}\Big(-\frac{\Delta_{\min}}{2(\log|\mcY|+2)}\Big)\bigg)\bigg)\bigg\}\ ,
        \end{align}
    \end{enumerate}
    where $W_{-1}$ denotes the non-principal real branch of the Lambert-$W$ function.
\end{theorem}

From Theorem~\ref{theorem: sample complexity}, we observe that the sample complexity has two components: (1) a component that scales as $O(1/\varepsilon^2)$, and is attributed to the number of samples required for the convergence of the policy induced by parameter estimates to the policy induced by the ground truth, $\pi^\star$, and, (2) a second component, $N_{\rm MaxMin}$, which is a price we pay for the MaxMin objective, and quantifies the number of samples required to ensure that $\widehat u = u^\star$. Note that in the large-gap regime, the additional price scales as $O\Big(\frac{1}{\Delta_{\min}^4}\Big)$, and it may be a small price for large values of $\Delta_{\min}$. On the other hand, since $W_{-1}(-x)\approx\log x$ for small $x\in\R_+$, we notice that there is an $O\Big(\frac{1}{\Delta_{\min}^4}\Big)$ scaling in the small-gap regime, which may be substantial. While the first component in~\eqref{eq:theorem_sample_complexity} has been reported in the literature (see, e.g., \cite{pmlr-v235-xiong24a}) as the sample complexity of uniform-reward RLHF, the second component is attributed to the MaxMin objective, and is a novel observation, which, to the best of our knowledge, is the first sample complexity guarantee of the MaxMin-RLHF objective.

\section{Experiments}
\label{sec:experiments}

In this section, we evaluate the performance of the SharedRep-RLHF algorithm on various language tasks, comparing it against MaxMin-RLHF. Our empirical study is guided by two central questions: (1) Does SharedRep-RLHF improve group fairness over MaxMin-RLHF for tasks with low minority representation, as measured by the {\em average minority score}? (2) Does SharedRep-RLHF achieve a higher {\em win rate} on these tasks under the same conditions? We summarize the key empirical findings in this section, while deferring experimental setup details, additional results, and ablation studies to Appendix~\ref{appendix: experiments}.

\paragraph{Tasks.} In this section, we present two distinct language tasks: {\bf controlled sentiment analysis} and {\bf mathematical reasoning}. Experimental results for {\bf single-turn dialogue} are deferred to Appendix~\ref{appendix: experiments}.

\textit{Controlled sentiment analysis.} We use the IMDb dataset~\cite{maas2011learning} following~\citeauthor{chakraborty2024maxminrlhfalignmentdiversehuman}, and adopt the prompt construction methodology from~\cite{rafailov2023direct}: for each review in the training-split, the first $2$–$8$ tokens are extracted as the prompt. To simulate group-specific preferences, the dataset is randomly partitioned into {\em majority} and {\em minority} sub-populations. The majority group values conciseness, while the minority group favors responses that are both positive and concise. Each response is assigned a group-specific gold score. For the majority, this score is a normalized conciseness measure computed from the length of the response. For the minority, the score is a convex combination of sentiment and conciseness, with a $70$\% weight on sentiment -- predicted using \texttt{lvwerra/distilbert-imdb} -- and $30$\% on conciseness. Preference labels are generated by comparing gold scores within each response pair and assigning the ``chosen'' label to the higher-scoring response. Notably, conciseness serves as a shared preference across both groups, which SharedRep-RLHF is designed to exploit.

\begin{table*}[t]
\small
\centering
\begin{tabular}{cccccc}
\toprule
\multirow{2}{*}{\shortstack{Minority\\Proportion}} 
& \multicolumn{3}{c}{Mean Minority Score}
& \multicolumn{2}{c}{Minority Win Rate (\%)} \\
\cmidrule(lr){2-4} \cmidrule(lr){5-6}
& MaxMin & SharedRep & Gold
& MaxMin & SharedRep \\
\midrule
0.01 & \num{0.589612} $\pm$ \num{0.000616} & \textbf{\num{0.687198}} $\pm$ \num{0.008603} & \multirow{4}{*}{\num{0.8084136299465374} $\pm$ \num{0.001445409734884519}} & 21.38 & \textbf{25.26} \\
0.05 & \num{0.628} $\pm$ \num{0.000601} & \textbf{\num{0.689723}} $\pm$ \num{0.014984} & & 25.21 & \textbf{26.01} \\
0.1 & \num{0.647} $\pm$ \num{0.000590} & \textbf{\num{0.706462}} $\pm$ \num{0.012821} & & 28.31 & \textbf{31.80} \\
0.2 & \num{0.673} $\pm$ \num{0.000572} & \textbf{\num{0.724814}} $\pm$ \num{0.012303} & & 23.97 & \textbf{31.49} \\
\bottomrule
\end{tabular}
\caption{Comparison of mean score and win rate (\%) between MaxMin- and SharedRep-RLHF ($K=2$) across different minority proportions for IMDb.}
\label{tab:mean_score_winrate_IMDb}
\end{table*}

\begin{table*}[t]
\centering
\small
\begin{tabular}{cccccc}
\toprule
\multirow{2}{*}{\shortstack{Minority\\Proportion}} 
& \multicolumn{3}{c}{Mean Minority Score}
& \multicolumn{2}{c}{Minority Win Rate (\%)} \\
\cmidrule(lr){2-4} \cmidrule(lr){5-6}
& MaxMin & SharedRep &  Gold
& MaxMin & SharedRep \\
\midrule
0.01 &  \num{0.239025} $\pm$ \num{0.002723} & \textbf{\num{0.310037}} $\pm$ \num{0.002126} &\multirow{4}{*}{\num{0.41453692978445056
} $\pm$ \num{0.0032619289734337528}}&  19.92 & \textbf{29.72}\\
0.05 &  \num{0.136607} $\pm$ \num{0.001588} & \textbf{\num{0.287745}} $\pm$ \num{0.002123} & &  4.93 & \textbf{25.32}\\
0.10 &  \num{0.194189} $\pm$ \num{0.001730} & \textbf{\num{0.293033}} $\pm$ \num{0.002352} & &  11.43 & \textbf{28.90}\\
0.15 &  \textbf{\num{0.316723}} $\pm$ \num{0.002699}  & \num{0.277359} $\pm$ \num{0.002421}  & &  \textbf{30.52} & 24.92\\
0.20 &  \textbf{\num{0.296033}} $\pm$ \num{0.003889} & \num{0.269524} $\pm$ \num{0.002006} & &  \textbf{31.65} & 22.76\\
\bottomrule
\end{tabular}
\caption{Comparison of mean score and win rate (\%) between MaxMin- and SharedRep-RLHF ($K=16$) across different minority proportions for GSM8K.}
\label{tab:mean_score_winrate_GSM8K}
\end{table*}

\textit{Mathematical reasoning.} For this task, we use the GSM8K dataset~\cite{cobbe2021training}, which contains high-quality grade school math word problems requiring multi-step arithmetic reasoning. 
We construct prompts by appending each question to a fixed two-shot prefix designed to elicit chain-of-thought (CoT) reasoning. Model-generated responses are evaluated using both correctness and stylistic preferences. To simulate pluralistic preferences, we define two sub-populations: the {\em majority}, which values brevity and correctness, and the {\em minority}, which values correctness and Socratic-style elaborate reasoning. For each response, correctness is determined by extracting the final numeric answer and comparing it to the ground truth. A normalized length score and a normalized Socratic score -- derived from a reward model fine-tuned to detect Socratic reasoning -- are used to compute subpopulation-specific gold scores. More specifically, the majority group’s gold score is computed as a weighted combination of response accuracy ($20$\%) and conciseness ($80$\%), reflecting a preference for brief and correct answers. In contrast, the minority group’s gold score assigns $20$\% weight to correctness and $80$\% to the degree of elaboration and step-by-step reasoning, capturing a preference for detailed, Socratic-style explanations. Note that response accuracy serves as a shared trait that informs the preferences of both the majority and minority sub-populations.

\paragraph{Models.}
For the controlled sentiment analysis task, we use \texttt{lvwerra/gpt2-imdb} as the policy model -- a \texttt{GPT2} model fine-tuned on the IMDb training split for one epoch. For MaxMin-RLHF, we attach a regression head to the \texttt{openai-community/gpt2} backbone to serve as the reward model. For SharedRep-RLHF, we extend the same backbone with a linear layer followed by two regression heads to capture both {\em shared} and {\em group-specific} components. For the mathematical reasoning task, we use \texttt{Qwen/Qwen2.5-Math-1.5B} as the policy model. This choice eliminates the need for supervised fine-tuning on GSM8K, as the model has been reported to achieve strong performance ($76.8$\% accuracy) on the dataset using few-shot CoT prompting. For reward modeling, we use the \texttt{openai-community/gpt2-large} backbone. As the gold-reward model for the Socratic score, we train a GPT2-large model on the Socratic train-split of the GSM8K dataset until convergence. For reward-learning, a single regression head is added for MaxMin-RLHF per group, while SharedRep-RLHF uses a linear layer followed by two regression heads to capture shared and group-specific weights. 

\paragraph{Results.} For both tasks, we evaluate (i) the average minority score achieved by the MaxMin- and SharedRep-RLHF policies on the test sets, and (ii) the win-rate of the policies against the ``gold-reward'' skylines, which use the ground truth rewards to learn a MaxMin policy. 

\textit{Controlled sentiment analysis.} Table~\ref{tab:mean_score_winrate_IMDb} reports the mean minority scores and win rates for SharedRep-RLHF and MaxMin-RLHF across varying minority group proportions on the IMDb task, along with the gold reward skyline which assumes to know the ground truth. SharedRep-RLHF consistently outperforms MaxMin-RLHF in terms of minority group satisfaction, achieving higher average scores across all proportion levels. Notably, when the minority proportion is just $1$\%, SharedRep-RLHF delivers a substantial $16.5$\% improvement over MaxMin-RLHF in mean minority score (0.687 vs.\ 0.590), highlighting its effectiveness in extreme imbalance scenarios. In addition to stronger performance, SharedRep-RLHF exhibits greater robustness: as the minority proportion decreases from $20$\% to $1$\%, MaxMin-RLHF suffers a $12.33$\% relative drop in minority score, whereas SharedRep-RLHF degrades by only $5.53$\%. This indicates that SharedRep-RLHF better preserves minority utility under growing data imbalance. We also compare win rates -- the proportion of preference pairs in which a model’s response is preferred over the gold responses. SharedRep-RLHF outperforms MaxMin-RLHF at every proportion level, with gains ranging from $0.8$ to $7.5$ percentage points. This further confirms that SharedRep-RLHF yields responses that are more aligned with minority preferences, even when such preferences are sparsely represented. 

\textit{Mathematical reasoning.} Table~\ref{tab:mean_score_winrate_GSM8K} shows the performance of MaxMin-RLHF and SharedRep-RLHF on GSM8K across varying minority proportions, and how these compare to MaxMin-RLHF with gold rewards. SharedRep-RLHF outperforms MaxMin-RLHF in terms of mean minority score in $3$ out of $5$ settings, with especially large gains at low proportions: at $5$\% minority, the mean score improves by over $110.2$\% (0.288 vs.\ 0.137). In terms of minority win rate, SharedRep-RLHF again dominates, improving over MaxMin-RLHF by $9.8$ to $20.39$ percentage points. The most significant difference is at $5$\%, where SharedRep-RLHF increases the win rate by approximately five-fold ($25.32$\% vs.\ $4.93$\%). These results highlight SharedRep-RLHF’s ability to generalize minority-aligned generation under severe imbalance, leveraging shared traits across groups to improve estimation fidelity.

\section{Conclusions}

In this paper, we revisited the problem of RLHF with diverse preferences. We identified a crucial bottleneck of the state-of-the-art (SOTA)~\cite{chakraborty2024maxminrlhfalignmentdiversehuman} and demonstrated that MaxMin-RLHF, purported toward achieving group-fairness, suffers significantly under data imbalances. We proposed to leverage shared traits across groups to improve reward estimation, and consequently, achieve policy alignment that is more robust in terms of fairness under significant data imbalances across groups. We provided provable guarantees that SharedRep-RLHF, our proposed algorithm, outperforms MaxMin-RLHF, and characterized the sample cost of achieving group-fairness compared to uniform RLHF. Finally, we conducte comprehensive experiments on various language tasks, which showcase that SharedRep-RLHF outperforms the SOTA in almost all experiments, exhibiting the advantage of learning shared traits to improve reward estimation.


\newpage
\bibliography{references}

\newpage
\appendix
\onecolumn

\section*{Organization of the Appendix}

The appendix is organized into three main sections: (i) an overview of uniform-preference RLHF and a supplementary literature review (Appendix~\ref{appendix: canonical RLHF}), (ii) complete proofs of the analytical results presented in the main paper (Appendix~\ref{proof: est_con} to Appendix~\ref{proof: theorem: sample complexity}), and (iii) extended experimental results (Appendix~\ref{appendix: experiments}). The proofs are presented in the order they appear in the main text, beginning with Lemma~\ref{lemma: est_con_sharedrep} in Appendix~\ref{proof: est_con}, followed by Theorem~\ref{theorem: performance comparison} in Appendix~\ref{proof: theorem: performance comparison}, Lemma~\ref{lemma:equivalence_of_u} in Appendix~\ref{proof: lemma:equivalence_of_u}, and concluding with Theorem~\ref{theorem: sample complexity} in Appendix~\ref{proof: theorem: sample complexity}. Appendix~\ref{appendix: experiments} provides additional experimental details for Section~\ref{sec:experiments}, including ablation studies and extended results on single-turn dialogue. We conclude with implementation details for the SharedRep-RLHF reward model using \texttt{huggingface-trl}.

\newpage

\section{Canonical (Uniform Preference) RLHF}
\label{appendix: canonical RLHF}
In this paper, we focus on {\em offline} RLHF. We begin by comparing offline versus online RLHF, and subsequently, delineate the uniform-preference RLHF pipeline.
\paragraph{Offline versus online RLHF.} Depending on the data acquisition mechanism for the reward-modeling step, two distinct paradigms have emerged, namely, {\em offline} and {\em online} RLHF. In offline RLHF~\cite{ouyang2022training,bai2022training, stiennon2022learningsummarizehumanfeedback,liu2024aligning,christian2021alignment}, a set of prompts is sampled from an arbitrary distribution. Subsequently, for each prompt, the SFT model is queried to obtain a set of responses, and an annotator (possibly human, or a powerful LLM) is asked to rank these responses. We then learn a reward function from the preference dataset using a choice model, and subsequently, proceed to the RL fine-tuning step. This is the preferred RLHF paradigm in scenarios where annotators are not readily accessible for real-time feedback. On the other hand, when we have online access to annotators, we may adopt a more efficient, preferably an {\em on-policy} sampling approach to construct the dataset and estimate the reward function in an adaptive and sample-efficient fashion~\cite{xie2024exploratory,pang2024iterative,cen2024value,zhang2024self}. 

\paragraph{RLHF pipeline.}

In the canonical RLHF setting, the learner has access to a {\em preference dataset}, and alignment is a three-step approach stated as follows.
\begin{enumerate}
    \item \textbf{Supervised fine-tuning (SFT):} As the first step, a pre-trained language model (LM) is further trained on a high-quality dataset in a supervised fashion based on the downstream task of interest. Let us denote the LM post SFT by the reference model $\pi_{\rm ref}$. 
    \item \textbf{Reward-modeling:} The premise in RLHF is that the learner has access to a preference dataset denoted by $\mcD \triangleq \{ (\bx_i,\by_i,\by^\prime_i,z_i)\}_{i=1}^N$. Here, for each data $i\in[N]$, $\bx_i$ denotes a prompt, $\by_i,\by_i^\prime\sim\pi_{\rm ref}(\cdot\med\bx_i)$ denotes two responses generated from the reference model conditioned on the prompt $\bx_i$, and $z_i\in\{0,1\}$ denotes the annotator preference, where $z_i=1$ if the annotator prefers $\by_i$ as a response to prompt $\bx_i$, and $z_i=0$ otherwise. Canonically, preferences are modeled using the Bradley-Terry (BT) choice model~\cite{bradley1952rank}, which is stated as follows.
    \begin{align}
    \label{eq:BTL}
        \P\Big (z = 1 \med \bx,\by,\by^\prime\Big )\;\triangleq\; \frac{\exp\big(r^\star(\bx,\by)\big)}{\exp\big(r^\star(\bx,\by)\big) + \exp\big(r^\star(\bx,\by^\prime)\big)} = \sigma \Big( r^\star(\bx,\by) - r^\star(\bx,\by^\prime)\Big)\ .
    \end{align}
    In the reward-modeling step, the learner assumes a parametric representation of the reward function $r^\star$, usually modeled by fixing the SFT backbone, removing the logits and softmax layers, and adding a single neuron (or a multi-layer perceptron) whose output is a scalar reward. Under a uniform preference setting, let $\btheta^\star$ denote the ground truth representing the users' preferences. The learner optimizes the binary cross entropy loss to learn an estimate $\widehat\btheta$ of $\btheta^\star$. More specifically, the learner forms the estimate
    \begin{align}
        \widehat\btheta \;\triangleq\;  \argmin_{\btheta}\;&
        \bigg\{\sum\limits_{i\in[N]} z_i\log\left(\sigma\Big(r_{\btheta}(\bx_i,\by_i) - r_{\btheta}(\bx_i,\by^\prime_i)\Big)\right) + (1-z_i)\log\left(1 - \sigma\Big(r_{\btheta}(\bx_i,\by_i) - r_{\btheta}(\bx_i,\by_i^\prime)\Big)\right)\bigg\}.
    \end{align}
    \item \textbf{RL fine-tuning:} Once we have learned the reward model $r_{\widehat\btheta}$, we aim to find a policy that maximizes the average reward while staying sufficiently close to the reference SFT model $\pi_{\rm ref}$. More specifically, we use policy gradient algorithms like PPO in order to learn the optimal policy that maximizes the regularized value function,
    \begin{align}
        \pi_{\rm single\: reward}\; \in\;\argmax_\pi\;\E_{\bx\sim\rho, \by\sim\pi(\cdot\med \bx)} \Big[ r_{\widehat\btheta}(\bx,\by) - \beta D_{\sf KL}\big(\pi(\cdot\med \bx)\|\pi_{\rm ref}(\cdot\med\bx) \big)\Big]\ ,
    \end{align}
    where $\rho$ denotes a distribution over the space of prompts $\mcX$, $D_{\rm KL}$ denotes the Kullback-Leibler (KL) divergence measure, and $\beta\in\R_+$ denotes the KL-penalty. Furthermore, let $\mcY$ denote the space of responses.
\end{enumerate}
\section{Proof of Lemma~\ref{lemma: est_con_sharedrep}}
\label{proof: est_con}
For some $\mu_{\max}\in\R_+$ let us define the function $\ell: [-\mu_{\max},\mu_{\max}]\mapsto\R$ as
\begin{align*}
    \ell(\mu)\;\triangleq\;-z\log \sigma(\mu) - (1-z)\log \Big( 1-\sigma(\mu)\Big)\ ,
\end{align*}
where 
\begin{align*}
    \sigma(\mu)\;\triangleq\; \frac{1}{1+\e^{-\mu}}
\end{align*}
denotes the sigmoid function, and $z\in\{0,1\}$. From first principles, we obtain that for any $\mu,\mu_0\in {\rm int\;dom}(\ell)$, 
\begin{align}
\label{eq:est_con_1}
    \ell(\mu)\;=\; \ell(\mu_0) + \dot{\ell}(\mu_0)(\mu-\mu_0) + \displaystyle\int_0^1 (1-t)(\mu - \mu_0)^2\ddot{\ell}\Big( t\mu + (1-t)\mu_0\Big)\diff t\ .
\end{align}
The following results may be readily verified.
\begin{itemize}
    \item[(a)] $\dot{\ell}(\mu)\;=\;-\frac{\e^{-\mu}}{1+\e^{-\mu}}\:z + \frac{1}{1+\e^{-\mu}}\:(1-z)$ , and,
    \item[(b)] $\ddot{\ell}(\mu)\;=\; \frac{1}{\e^{\mu} + \e^{-\mu} + 2}$ .
\end{itemize}
Furthermore, from the assumption that $\mu\in[-\mu_{\max},\mu_{\max}]$, we have
\begin{align}
\label{eq:est_con_2}
    \ddot{\ell}(\mu)\;\geq\;\underbrace{\frac{1}{\e^{\mu_{\max}} + \e^{-\mu_{\max}} + 2}}_{{\triangleq\; 2\gamma}}\ .
\end{align}
Combining~\eqref{eq:est_con_1} and~\eqref{eq:est_con_2}, we obtain that
\begin{align}
    \ell(\mu)\;&\geq\; \ell(\mu_0) + \dot{\ell}(\mu_0)(\mu-\mu_0) + 2\gamma(\mu-\mu_0)^2\displaystyle\int_0^1 (1-t)\diff t\nonumber\\
    &=\;\ell(\mu_0) + \dot{\ell}(\mu_0)(\mu-\mu_0) + \gamma(\mu-\mu_0)^2\ .
    \label{eq:est_con_3}
\end{align}
Recall the definition of the (empirical) cross-entropy loss with linear function approximation,
\begin{align*}
    \mcL(\bB,\bW)\;=\; -\frac{1}{N} \sum\limits_{u\in[U]} \sum\limits_{i\in\mcH_u} z_i\log\Big( \sigma\big(\big\langle\bdelta_i,\bB\bw_u\big\rangle\big)\Big) + (1-z_i)\log\Big( 1-\sigma\big(\big\langle \bdelta_i,\bB\bw_u\big\rangle\big)\Big)\ ,
\end{align*}
where we have defined
\begin{align*}
    \bdelta_i\;\triangleq\; \phi(\bx_i,\by_i) - \phi(\bx_i,\by_i^\prime)\ .
\end{align*}
For any sub-population $u\in[U]$ and datapoint annotated by a human from the sub-population $i\in\mcH_u$, let us define
\begin{align*}
    \ell_i(\bB\bw_u)\;\triangleq\; -z_i\log\Big( \sigma\big(\big\langle\bdelta_i,\bB\bw_u\big\rangle\big)\Big) - (1-z_i)\log\Big( 1-\sigma\big(\big\langle \bdelta_i,\bB\bw_u\big\rangle\big)\Big)\ .
\end{align*}
Next, in~\eqref{eq:est_con_3} let us make the following substitutions.
\begin{itemize}
    \item $\ell = \ell_i$
    \item $\mu_0 = \big\langle \bdelta_i,\bB^\star\bw_u^\star\big\rangle$, and,
    \item $\mu = \big\langle \bdelta_i,\widehat\bB\widehat\bw_u\big\rangle$.
\end{itemize}
Furthermore, note that 
\begin{align*}
    \big\langle \bdelta_i,\bB^\star\bw_u^\star\big\rangle \;&\leq\;\| \bdelta_i\|_2 \| \bB^\star\bw_u^\star\|_2\\
    &=\; \| \phi(\bx_i,\by_i) - \phi(\bx_i,\by_i^\prime)\|_2\cdot\Big\| \sum\limits_{i\in[K]}\bB^\star(:,i)w_{u}^\star(i)\Big\|_2\\
    &\leq 2L_{\max} \Big(\sum\limits_{i\in[K]} w^\star_{u}(i)\| \bB^\star(:,i)\|_2\Big)\\
    &\leq 2L_{\max}B_{\max}\ .
\end{align*}
Hence, we make the substitution $\mu_{\max} = 2L_{\max}B_{\max}$. After making all the substitutions into~\eqref{eq:est_con_3}, we have
\begin{align*}
    \ell_i(\widehat B \widehat\bw_u) - \ell_i(\bB^\star \bw_u^\star) - \dot{\ell_i}(\bB^\star\bw_u^\star)\Big( \big\langle \bdelta_i, \widehat\bB\widehat\bw_u - \bB^\star\bw_u^\star\big\rangle\Big)\;\geq\;\gamma\Big( \big\langle \bdelta_i, \widehat\bB\widehat\bw_u - \bB^\star\bw_u^\star\big\rangle\Big)^2\ .
\end{align*}
Averaging over all datapoints, we have
\begin{align}
    \mcL(\widehat\bB, \widehat\bW) - \mcL(\bB^\star,\bW^\star) &- \frac{1}{N}\sum\limits_{u\in[U]}\sum\limits_{i\in\mcH_u} \dot{\ell_i}(\bB^\star\bw_u^\star)\Big( \big\langle \bdelta_i, \widehat\bB\widehat\bw_u - \bB^\star\bw_u^\star\big\rangle\Big)\nonumber\\
    &\geq\; \frac{\gamma}{N}\sum\limits_{u\in[U]}\sum\limits_{i\in\mcH_u} \Big( \big\langle \bdelta_i, \widehat\bB\widehat\bw_u - \bB^\star\bw_u^\star\big\rangle\Big)^2\ .
    \label{eq:est_con_4}
\end{align}
Furthermore, from the definition of MLE, we have
\begin{align}
    \mcL(\widehat\bB, \widehat\bW) - &\mcL(\bB^\star,\bW^\star) - \frac{1}{N}\sum\limits_{u\in[U]}\sum\limits_{i\in\mcH_u} \dot{\ell_i}(\bB^\star\bw_u^\star)\Big( \big\langle \bdelta_i, \widehat\bB\widehat\bw_u - \bB^\star\bw_u^\star\big\rangle\Big)\;\nonumber\\
    &\leq\;-\frac{1}{N}\sum\limits_{u\in[U]}\sum\limits_{i\in\mcH_u} \dot{\ell_i}(\bB^\star\bw_u^\star)\Big( \big\langle \bdelta_i, \widehat\bB\widehat\bw_u - \bB^\star\bw_u^\star\big\rangle\Big)\ .
    \label{eq:est_con_5}
\end{align}
Next, defining
\begin{align*}
    V_i(u)\;\triangleq\;
    \begin{cases}
        -\frac{\e^{-\big\langle \bdelta_i,\bB^\star\bw_u^\star\big\rangle}}{1 + \e^{-\big\langle \bdelta_i,\bB^\star\bw_u^\star\big\rangle}}, \qquad&\text{w.p.}\qquad \sigma\Big( \big\langle \bdelta_i,\bB^\star\bw_u^\star\big\rangle\Big),\\
        \frac{1}{1 + \e^{-\big\langle \bdelta_i,\bB^\star\bw_u^\star\big\rangle}}, \qquad&\text{w.p.}\qquad 1-\sigma\Big( \big\langle \bdelta_i,\bB^\star\bw_u^\star\big\rangle\Big) .
    \end{cases}\ ,
\end{align*}
it is easy to see that
\begin{align*}
    -\frac{1}{N}\sum\limits_{u\in[U]}\sum\limits_{i\in\mcH_u} \dot{\ell_i}(\bB^\star\bw_u^\star) \Big( \big\langle \bdelta_i, \widehat\bB\widehat\bw_u - \bB^\star\bw_u^\star\big\rangle\Big)  &\stackrel{(a)}{=} -\frac{1}{N} \sum\limits_{u\in[U]}\sum\limits_{i\in\mcH_u}V_i(u)\Big( \big\langle \bdelta_i, \underbrace{\widehat\bB\widehat\bw_u - \bB^\star\bw_u^\star}_{{\triangleq\;\bDelta(u)}}\big\rangle\Big)\nonumber \\
    \quad &= \Big\langle -\frac{1}{N}\bX^\top\bv, \bDelta(u)\Big\rangle\ ,
\end{align*}
where, we have defined the vector $\bv\triangleq [v_1(1),\cdots,v_{|\mcH_U|}(U)]^\top$, and $\bX\in\R^{N\times d}$ is defined such that its rows are the difference vectors $\{\bdelta_i : i\in[N]\}$. Applying Cauchy-Schwarz inequality, we further obtain that
\begin{align}
    -\frac{1}{N}\sum\limits_{u\in[U]}\sum\limits_{i\in\mcH_u} \dot{\ell_i}(\bB^\star\bw_u^\star) \Big( \big\langle \bdelta_i, \widehat\bB\widehat\bw_u - \bB^\star\bw_u^\star\big\rangle\Big)\;\leq\;\Big\| \frac{1}{N} \bX^\top\bv\Big\|_{\Big(\frac{1}{N}\bX^\top\bX + \lambda\bI \Big)^{-1}}\cdot \|\bDelta(u) \|_{\Big(\frac{1}{N}\bX^\top\bX + \lambda\mathbb{I} \Big)}\ .
    \label{eq:est_con_6}
\end{align}
In order to upper-bound the first term in~\eqref{eq:est_con_6}, we observe that
\begin{align*}
    \Big\| \frac{1}{N} \bX^\top\bv\Big\|^2_{\Big(\frac{1}{N}\bX^\top\bX + \lambda\mathbb{I} \Big)^{-1}} \;&=\; \bv^\top\underbrace{\left( \frac{1}{N^2}\bX\left( \frac{1}{N}\bX^\top\bX + \lambda\mathbb{I}\right)^{-1}\bX^\top\right)}_{{\triangleq\; \bM}}\bv\ .
\end{align*}
Noting that $\bv$ is a binary random vector, we adopt the exact line of arguments as~\cite[Lemma 3.1]{zhu2023principled}, and for any $\delta\in(0,1)$, with probability at least $(1-\delta)$, we obtain the following upper-bound for some universal constant $C_1\in\R_+$.
\begin{align}
    \Big\| \frac{1}{N} \bX^\top\bv\Big\|^2_{\Big(\frac{1}{N}\bX^\top\bX + \lambda\mathbb{I} \Big)^{-1}}\;\leq\; C_1\cdot\frac{d+\log(1/\delta)}{N}\ .
    \label{eq:est_con_7}
\end{align}
Combining~\eqref{eq:est_con_4}, \eqref{eq:est_con_5}, \eqref{eq:est_con_6}, and~\eqref{eq:est_con_7}, with probability at least $1-\delta$, we have
\begin{align*}
    \gamma\|\bDelta(u) \|^2_{\Big( \frac{1}{N}\bX^\top\bX\Big)}\;\leq\; C_1\cdot\frac{d+\log(1/\delta)}{N}\cdot\| \bDelta(u)\|_{\Big( \frac{1}{N}\bX^\top\bX + \lambda\mathbb{I}\Big)}\ .
\end{align*}
Rearranging, it can be readily verified that with probability at least $(1-\delta)$, and for some universal constant $C\in\R_+$ we have
\begin{align*}
    \Big\| \widehat\bB\widehat\bw_u - \bB^\star\bw_u^\star\Big\|_{\Big( \frac{1}{N}\bX^\top\bX + \lambda\mathbb{I}\Big)}\;\leq C\sqrt{\frac{d + \log(1/\delta)}{N} + \lambda B_{\max}^2}\ ,
\end{align*}
which proves our claim.
\section{Proof of Theorem~\ref{theorem: performance comparison}}
\label{proof: theorem: performance comparison}

Recall the definition of the optimal policy $\pi^\star_u$ that maximizes the value function $J_u(\pi)$ defined in~\eqref{eq:GT_unregularized_value_function}. It can be readily verified that
\begin{align}
    \pi^\star_u\;\in\;\argmax_\pi\;\sum\limits_{\bx\in\mcX}\underbrace{\left(\sum\limits_{\by\in\mcY} r_{\btheta^\star}(\bx,\by)\pi(\by\med\bx)\right)}_{(*)}\rho(\bx)\ .
\end{align}
Furthermore, note that the optimal policy $\pi^\star_u$ is the one that maximizes $(*)$ for every prompt $\bx\in\mcX$. This is achieved by the deterministic policy that chooses the response
\begin{align}
    \pi_u^\star(\bx)\;\;=\;\argmax_{\by\in\mcY}\; r_{\btheta^\star_u}(\bx,\by)\ ,
\end{align}
given any prompt $\bx$, where we write $\pi_u^\star(\bx)$ with slight abuse of notation, denoting the response that $\pi^\star_u$ chooses for $\bx$. Similarly, by additionally invoking Assumption~\ref{assumption:linear reward}, we obtain
\begin{align}
\label{eq:pi(x)}
    \widehat\pi_u^{\rm SR}(\bx)\;&=\;\argmax_\pi\;\min_{(\bB,\bw)\in\Theta_{\rm SR}(\widehat\bB,\widehat\bw)}\;\big\langle \phi(\bx,\pi(\bx)),\bB\bw_u\big\rangle\ ,\nonumber\\
    \text{and}\quad \widehat\pi_u^{\rm MM}(\bx)\;&=\;\argmax_\pi\;\min_{\btheta\in\Theta_{\rm MM}(\widehat\btheta_u)}\;\big\langle \phi(\bx,\pi(\bx)),\btheta\big\rangle\ ,
\end{align}
where $\pi(\bx)$ denotes a deterministic policy $\pi: \mcX\mapsto\mcY$, and $\widehat\pi_u^{\rm SR}(\bx)$ and $\widehat\pi_{u}^{\rm MM}(\bx)$ denote the responses induced by the MaxMin- and {SharedRep}-RLHF estimators, respectively, for a given prompt $\bx$. Furthermore, note that minimizers of the inner optimization problems in~\eqref{eq:pi(x)} lie at the boundary of the constraint sets, since minimum over linear functions is a concave function. Accordingly, it can be readily verified (see, e.g., \cite[Chapter $19$]{lattimore2020bandit}) that
\begin{align}
    \widehat\pi_u^{\rm SR}(\bx)\;&\in\; \argmax_{\by\in\mcY}\;\big\langle \phi(\bx,\by),\widehat\bB\widehat\bw_u\big\rangle - \eta^{\rm SR}(N,\lambda,\delta)\|\phi(\bx,\by) \|_{\big( \Sigma + \lambda\mathbb{I}\big)^{-1}}\ ,\nonumber\\
    \text{and}\quad \widehat\pi_u^{\rm MM}(\bx)\;&\in\; \argmax_{\by\in\mcY}\;\big\langle \phi(\bx,\by),\widehat\btheta_u\big\rangle - \eta^{\rm MM}(N_u,\lambda,\delta)\|\phi(\bx,\by) \|_{\big( \Sigma_u + \lambda\mathbb{I}\big)^{-1}}\ ,
\end{align}
where $\eta^{\rm SR}$ has been defined in Algorithm~\ref{algorithm:SharedRep} and we define 
\begin{align}
    \eta^{\rm MM}(N_u,\lambda,\delta)\;\triangleq\;C_{\rm MM}\sqrt{\frac{d + \log(1/\delta)}{\gamma^2N_u}+\lambda B_{\max}^2}\ .
\end{align}
Next, with probability at least $1-\delta$, we have
\begin{align}
    {\rm SubOpt}_u(\widehat\pi_u^{\rm MM}) - {\rm SubOpt}_u(\widehat\pi_u^{\rm SR})\;&=\; J_u(\widehat\pi_u^{\rm SR}) - J_u(\widehat\pi_u^{\rm MM})\\
    &\geq\; \widehat J_u^{\rm SR}(\widehat\pi_u^{\rm SR}) - J_u(\widehat\pi_u^{\rm MM})\\
    &=\;\underbrace{\widehat J_u^{\rm SR}(\widehat\pi_u^{\rm SR}) - \widehat J_u^{\rm SR}(\widehat\pi_u^{\rm MM})}_{\triangleq\;A_1} + \underbrace{\widehat J_u^{\rm SR}(\widehat\pi_u^{\rm MM}) - J_u(\widehat\pi_u^{\rm MM})}_{\triangleq\;A_2}\ . 
\end{align}
\underline{\textbf{Lower bounding $A_2$:}} $A_2$ captures the value function estimation error under the policy $\widehat\pi_u^{\rm MM}$. We will lower bound $A_2$ by upper bounding $-A_2$ as follows.
\begin{align}
    -A_2\;&=\; \E_{\bx\sim\rho}\big[ r_{\btheta^\star_u}(\bx,\widehat\pi_u^{\rm MM}(\bx))\big] - \underbrace{\min_{(\bB,\bw)\in\Theta(\widehat\bB,\widehat\bw_u)}\E_{\bx\sim\rho}\big[ r_{\bB\bw}(\bx,\widehat\pi_u^{\rm MM}(\bx))\big]}_{(**)}\\
    \label{eq:aligngap_1}
    &=\;\E_{\bx\sim\rho}\Big[\big\langle \phi(\bx,\widehat\pi_u^{\rm MM}(\bx))\:,\:\btheta_u^\star - \bB^\prime\bw_u^\prime\big\rangle \Big]\\
    \label{eq:aligngap_2}
    &\leq\;\E_{\bx\sim\rho}\Big[ \|\phi(\bx,\widehat\pi_u^{\rm MM}(\bx)) \|_{(\Sigma + \lambda\mathbb{I})^{-1}}\cdot\| \btheta^\star_u - \bB^\prime\bw_u^\prime\|_{\Sigma + \lambda\mathbb{I}}\Big]\\
    \label{eq:aligngap_3}
    &\leq\;2\eta^{\rm SR}(N,\lambda,\delta)\cdot\E_{\bx\sim\rho}\Big[\|\phi(\bx,\widehat\pi_u^{\rm MM}(\bx)) \|_{(\Sigma + \lambda\mathbb{I})^{-1}}\Big]\ ,
\end{align}
where~\eqref{eq:aligngap_1} follows from Assumption~\ref{assumption:linear reward} and denoting the minimizer in $(**)$ by $(\bB^\prime,\bw_u^\prime)$, \eqref{eq:aligngap_2} follows from Cauchy-Schwarz inequality, an~\eqref{eq:aligngap_3} follows from Lemma~\ref{lemma: est_con_sharedrep}. 

\noindent\underline{\textbf{Lower bounding $A_1$:}} $A_1$ captures the policy estimation error under the estimated value function in the {SharedRep}-RLHF framework. We have
\begin{align}
    A_1\;&=\; \underbrace{\min_{(\bB,\bw)\in\Theta(\widehat\bB,\widehat\bw_u)}\E_{\bx\sim\rho}\big[ r_{\bB\bw}(\bx,\widehat\pi_u^{\rm SR}(\bx))\big]}_{(\dagger)} - \E_{\bx\sim\rho}\big[ r_{\bB^\prime\bw_u^\prime}(\bx,\widehat\pi_u^{\rm MM}(\bx))\big]\\
    \label{eq:aligngap_4}
    &\geq\;\E_{\bx\sim\rho}\Big[ \big|\big\langle \phi(\bx,\widehat\pi_u^{\rm SR}(\bx)) - \phi(\bx,\widehat\pi_u^{\rm MM}(\bx)), \widetilde\bB\widetilde\bw_u\big\rangle\big |\Big]\\
    &\geq\;\inf_{\bB\in\mcB,\bw\in\Delta^{K-1}}\E_{\bx\sim\rho}\Big[  \big|\big\langle \phi(\bx,\widehat\pi_u^{\rm SR}(\bx)) - \phi(\bx,\widehat\pi_u^{\rm MM}(\bx)), \bB\bw\big\rangle\big |\Big] \\
    &\geq\;\inf_{\pi\neq\pi^\prime}\inf_{\bB\in\mcB,\bw\in\Delta^{K-1}} \E_{\bx\sim\rho}\Big[\big|\big\langle \phi(\bx,\pi(\bx)) - \phi(\bx,\pi^\prime(\bx)), \bB\bw\big\rangle\big |\Big]\\
    &=\;\inf_{\pi\neq\pi^\prime} \inf_{\bB\in\mcB,\bw\in\Delta^{K-1}} \sum\limits_{\bx\in\mcX} \big | \big\langle \phi(\bx,\pi(\bx)) - \phi(\bx,\pi^\prime(\bx)), \bB\bw\big\rangle\big |\rho(\bx)\\
    &\geq\;\inf_{\pi\neq\pi^\prime} \inf_{\bB\in\mcB,\bw\in\Delta^{K-1}} \rho_{\min}\sum\limits_{\bx\in\mcX} \big | \big\langle \phi(\bx,\pi(\bx)) - \phi(\bx,\pi^\prime(\bx)), \bB\bw\big\rangle\big |\\
    &\geq\;\inf_{\pi\neq\pi^\prime} \inf_{\bB\in\mcB,\bw\in\Delta^{K-1}} \max_{\bx\in\mcX}\;\rho_{\min} \big | \big\langle \phi(\bx,\pi(\bx)) - \phi(\bx,\pi^\prime(\bx)), \bB\bw\big\rangle\big |\\
    &\geq\;\inf_{\by\neq\by^\prime} \inf_{\bB\in\mcB,\bw\in\Delta^{K-1}} \max_{\bx\in\mcX}\;\rho_{\min} \big | \big\langle \phi(\bx,\by) - \phi(\bx,\by^\prime), \bB\bw\big\rangle\big |\\
    &=\;\rho_{\min}\xi_u\ ,
    \label{eq:aligngap_5}
\end{align}
where~\eqref{eq:aligngap_4} follows from Assumption~\ref{assumption:linear reward}, defining $(\widetilde\bB,\widetilde\bw_u)$ as the minimizer in $(\dagger)$, and the absolute value can be imposed since $A_1$ is always positive. The proof is concluded by combining~\eqref{eq:aligngap_3} and~\eqref{eq:aligngap_5}.
\section{Proof of Lemma~\ref{lemma:equivalence_of_u}}
\label{proof: lemma:equivalence_of_u}

Consider the set of reward functions $\{r_u : u\in[U]\}$, and recall the definition of the worst-case subpopulation $u_r$ from~\eqref{eq:u_r} and $\pi_r$ as defined in~\eqref{eq:pi_r}. We have,
\begin{align}
    \pi_r(\cdot\med\bx)\;&\in\; \argmax_{\pi}\min_u \;\E_{\bx\sim\rho,\by\sim\pi(\cdot\med\bx)}\left[r_u(\bx,\by) - \beta\log\frac{\pi(\by\med\bx)}{\pi_{\rm ref}(\by\med\bx)}\right]\\
    &=\; \argmin_\pi \max_u\; \E_{\bx\sim\rho,\by\sim\pi(\cdot\med\bx)}\left[ -\frac{1}{\beta}r_u(\bx,\by) -\log\frac{\pi_{\rm ref}(\by\med\bx)}{\pi(\by\med\bx)}\right]\\
    &=\;\argmin_\pi\max_u\;\E_{\bx\sim\rho,\by\sim\pi(\cdot\med\bx)}\left[\ln\left( \exp\Big( -\frac{1}{\beta}r_u(\bx,\by)\Big)\frac{\pi(\by\med\bx)}{\pi_{\rm ref}(\by\med\bx)}\right)\right]\\
    &=\;\argmin_\pi\max_u\;\E_{\bx\sim\rho,\by\sim\pi(\cdot\med\bx)}\bigg[\ln\left( \frac{\sum_{\by}\pi_{\rm ref}(\by\med\bx)\exp\Big( \frac{1}{\beta}r_u(\bx,\by)\Big)}{\pi_{\rm ref}(\by\med\bx)\exp\Big(\frac{1}{\beta}r_u(\bx,\by)\Big)}\cdot\pi(\by\med\bx)\right) \nonumber\\
    &\qquad\qquad\qquad\qquad\qquad\qquad-\ln \Big( \sum_{\by}\pi_{\rm ref}(\by\med\bx)\exp\Big( \frac{1}{\beta}r_u(\bx,\by)\Big)\Big)\bigg]\\
    &=\;\argmin_{\pi}\underbrace{\max_u\;\E_{\bx\sim\rho,\by\sim\pi(\cdot\med\bx)}\left[\ln\frac{\pi(\by\med\bx)}{\nu_{r_u}(\by\med\bx)} \right]}_{(\ast)}\ ,
    \label{eq:u_lemma_1}
\end{align}
where, in~\eqref{eq:u_lemma_1}, we recall the definition of $\nu_r$ from~\eqref{eq:Gibbs_general}. Furthermore, noting that $u_r$ is also the maximizer of $(\ast)$, we have,
\begin{align}
    u_r\;&\in\;\argmax_u\;\E_{\bx\sim\rho,\by\sim\pi(\cdot\med\bx)}\left[\ln\frac{\pi_r(\by\med\bx)}{\nu_{r_u}(\by\med\bx)} \right]\\
    &=\;\argmax_u\;\E_{\bx\sim\rho,\by\sim\pi(\cdot\med\bx)}\left[ \ln\frac{1}{\nu_{r_u}(\by\med\bx)}\right]\\
    &=\;\argmax_u\; \E_{\bx\sim\rho}\Big[ H\big(\nu_{r_u} (\cdot\med\bx)\big)\Big]\\
    &=\;\argmax_u\; H\big( \nu_{r_u}(\cdot\med\bX)\big)\ .
\end{align}

\section{Proof of Theorem~\ref{theorem: sample complexity}}
\label{proof: theorem: sample complexity}

We have
\begin{align}
    J_{\rm MaxMin}&(\pi^\star) - J_{\rm MaxMin}(\widetilde\pi^{\rm SR})\nonumber\\
    &=\;\underbrace{\E_{\bx\sim\rho,\by\sim\pi^\star(\cdot\med\bx)}\left[r_{\btheta^\star_{u^\star}}(\bx,\by) - \beta\log\frac{\pi^\star(\by\med\bx)}{\pi_{\rm ref}(\by\med\bx)}\right]}_{\triangleq\;J^\star}\nonumber\\
    &\qquad- \E_{\bx\sim\rho,\by\sim\widetilde\pi^{\rm SR}(\cdot\med\bx)}\left[r_{\btheta^\star_{\widehat u}}(\bx,\by) - \beta\log\frac{\widetilde\pi^{\rm SR}(\by\med\bx)}{\pi_{\rm ref}(\by\med\bx)}\right]\\
    &=\;J^\star - \E_{\bx\sim\rho,\by\sim\widetilde\pi^{\rm SR}(\cdot\med\bx)}\left[\left(r_{\btheta^\star_{\widehat u}}(\bx,\by) - \beta\log\frac{\widetilde\pi^{\rm SR}(\by\med\bx)}{\pi_{\rm ref}(\by\med\bx)}\right)\Big(\mathds{1}_{\{\widehat u = u^\star\}} + \mathds{1}_{\{\widehat u \neq u^\star\}}\Big)\right]\\
    &=\; \underbrace{\left(J^\star - \E_{\bx\sim\rho,\by\sim\widetilde\pi^{\rm SR}(\cdot\med\bx)}\left[ r_{\btheta^\star_{\widehat u}}(\bx,\by) - \beta\log\frac{\widetilde\pi^{\rm SR}(\by\med\bx)}{\pi_{\rm ref}(\by\med\bx)}\:\Big |\:\widehat u = u^\star\right]\right)}_{\triangleq\;A_3}\P(\widehat u = u^\star)\nonumber\\
    &\qquad+\;\left(J^\star - \E_{\bx\sim\rho,\by\sim\widetilde\pi^{\rm SR}(\cdot\med\bx)}\left[ r_{\btheta^\star_{\widehat u}}(\bx,\by) - \beta\log\frac{\widetilde\pi^{\rm SR}(\by\med\bx)}{\pi_{\rm ref}(\by\med\bx)}\:\Big |\:\widehat u \neq u^\star\right]\right)\P(\widehat u \neq u^\star)\ .
\end{align}
Using the same analysis as~\cite[Theorem $1$]{pmlr-v235-xiong24a}, it can be readily verified that
\begin{align}
\label{eq:A_3}
    A_3\;\leq 2\eta^{\rm SR}(N,\lambda,\delta)\big\| \E_{\bx\sim\rho,\by\sim\pi^\star(\cdot\med\bx)}[\phi(\bx,\by)]\big\|_{(\Sigma + \lambda\mathbb{I})^{-1}}\ .
\end{align}
Next, we will analyze the probability of misidentifying the worst-case subpopulation, i.e., $\P(\widehat u \ne u^\star)$. For quantifying $\P(\widehat u \ne u^\star)$, we will use a Fannes'-type inequality for upper-bounding the difference between the entropy of the estimated Gibbs distribution and the ground truth Gibbs distribution. Specifically, we leverage the following result.
\begin{lemma}[Theorem $1$, \cite{audenaert2006sharp}]
\label{lemma:Fannes}
    Let $\F$ and $\G$ denote finite-dimensional distributions supported on $\Omega$. We have
    \begin{align}
        H(\F) - H(\G)\;\leq\; D_{\rm TV}(\F\|\G)\cdot\log|\Omega| + h\big(D_{\rm TV}(\F\|\G)\big)\ , 
    \end{align}
    where $h(\cdot)$ denotes the binary entropy function.
\end{lemma}
We begin by bounding the conditional entropy of the Gibbs distributions induced by the true reward models and the estimates formed by the SharedRep-RLHF algorithm. Recall that $\{\nu_u^\star : u\in[U]\}$ denotes the set of Gibbs distributions induced by the true reward model. Furthermore, let us denote $\{\widetilde\nu_u: u\in[U]\}$ to be the set of Gibbs distributions induced by plugging the estimated rewards $\{r_{\widehat\bB\widehat\bw_u} : u\in[U]\}$ in~\eqref{eq:Gibbs_general}. For any $u\in[U]$ we have
\begin{align}
    &\Big| H\big( \nu^\star_u(\cdot\med\bX)\big) - H\big( \widetilde\nu_u(\cdot\med\bX)\big)\Big|\;\nonumber\\
    &=\;\Big| \E_{\bx\sim\rho}\big[H\big( \nu^\star_u(\cdot\med\bx)\big) - H\big( \widetilde\nu_u(\cdot\med\bx)\big)\big]\Big|\\
    \label{eq:sample_complexity_1}
    &\leq\; \E_{\bx\sim\rho}\left[ \Big| H\big( \nu^\star_u(\cdot\med\bx)\big) - H\big( \widetilde\nu_u(\cdot\med\bx)\big)\Big|\right]\\
    \label{eq:sample_complexity_2}
    &\leq\;\E_{\bx\sim\rho}\left[D_{\rm TV}\big(\nu_u^\star(\cdot\med\bx)\|\widetilde\nu_u(\cdot\med\bx) \big)\log|\mcY| + h\big(D_{\rm TV}\big(\nu_u^\star(\cdot\med\bx)\|\widetilde\nu_u(\cdot\med\bx) \big)\big)\right]\\
    \label{eq:sample_complexity_3}
    &\leq \E_{\bx\sim\rho}\left[ \sqrt{2D_{\rm KL}\big(\nu_u^\star(\cdot\med\bx)\|\widetilde\nu_u(\cdot\med\bx) \big)}\log|\mcY| + h\big( \sqrt{2D_{\rm KL}\big(\nu_u^\star(\cdot\med\bx)\|\widetilde\nu_u(\cdot\med\bx) \big)}\big)\right]\\
    \label{eq:sample_complexity_3_1}
    &<\;\E_{\bx\sim\rho}\bigg[\sqrt{2D_{\rm KL}\big(\nu_u^\star(\cdot\med\bx)\|\widetilde\nu_u(\cdot\med\bx) \big)}\Big(1-\log \sqrt{2D_{\rm KL}\big(\nu_u^\star(\cdot\med\bx)\|\widetilde\nu_u(\cdot\med\bx) \big)}\Big)\nonumber\\
    &\qquad\qquad\qquad\qquad+\sqrt{2D_{\rm KL}\big(\nu_u^\star(\cdot\med\bx)\|\widetilde\nu_u(\cdot\med\bx) \big)}\log|\mcY|\bigg]\\
    &=\;\E_{\bx\sim\rho}\bigg[ - \sqrt{2D_{\rm KL}\big(\nu_u^\star(\cdot\med\bx)\|\widetilde\nu_u(\cdot\med\bx) \big)}\log\sqrt{2D_{\rm KL}\big(\nu_u^\star(\cdot\med\bx)\|\widetilde\nu_u(\cdot\med\bx) \big)}\nonumber\\
    &\qquad\qquad\qquad\qquad \sqrt{2D_{\rm KL}\big(\nu_u^\star(\cdot\med\bx)\|\widetilde\nu_u(\cdot\med\bx) \big)}\big(\log|\mcY|+1\big)\bigg]\ ,
\end{align}
where~\eqref{eq:sample_complexity_1} follows from Jensen's inequality, \eqref{eq:sample_complexity_2} follows from Lemma~\ref{lemma:Fannes}, ~\eqref{eq:sample_complexity_3} follows from Pinsker's inequality, and~\eqref{eq:sample_complexity_3_1} follows from the fact that $h(p)< p(1-\log p)$. Subsequently, we have the following two cases.
\begin{itemize}
    \item \textbf{Case $1$ ($D_{\rm KL}\big(\nu_u^\star(\cdot\med\bx)\|\widetilde\nu_u(\cdot\med\bx) \big)\geq\frac{1}{2\e^2}$):} In this case, it can be readily verified that
    \begin{align}
        \Big| H\big( \nu^\star_u(\cdot\med\bX)\big) - H\big( \widetilde\nu_u(\cdot\med\bX)\big)\Big|\;\leq\; \E_{\bx\sim\rho}\Big[\sqrt{2D_{\rm KL}\big(\nu_u^\star(\cdot\med\bx)\|\widetilde\nu_u(\cdot\med\bx) \big)}\Big] \big(\log|\mcY| + 2\big)\ .
    \end{align}
    \item \textbf{Case $2$ ($D_{\rm KL}\big(\nu_u^\star(\cdot\med\bx)\|\widetilde\nu_u(\cdot\med\bx) \big)<\frac{1}{2\e^2}$):} In this case, it can be readily verified that
    \begin{align}
        \Big| H\big( \nu^\star_u(\cdot\med\bX)\big) - H\big( \widetilde\nu_u(\cdot\med\bX)\big)\Big|\;&\leq\; \E_{\bx\sim\rho}\Big[ -\sqrt{2D_{\rm KL}\big(\nu_u^\star(\cdot\med\bx)\|\widetilde\nu_u(\cdot\med\bx) \big)}\nonumber\\
        &\quad\times\log \sqrt{2D_{\rm KL}\big(\nu_u^\star(\cdot\med\bx)\|\widetilde\nu_u(\cdot\med\bx) \big)}\Big]\big(\log|\mcY| + 2\big)\ .
    \end{align}
\end{itemize}
Based on the cases $1$ and $2$, we have obtained an upper bound $|H(\nu_u^\star(\cdot\med\bX)) - H(\widetilde\nu_u(\cdot\med\bX))| < \E_{\bx\sim\rho}[f(D_{\rm KL}(\nu^\star(\cdot\med\bx)\|\widetilde\nu_u(\cdot\med\bx)))]$, where we define the function $f$ as
\begin{align}
\label{eq:f}
f(x) \triangleq
    \begin{cases}
    \sqrt{2x}(\log|\mcY|+2), \qquad \qquad \qquad &\text{if} \; x\geq\frac{1}{2\e^2} \\
    &\\
    -(\log|\mcY|+2)\sqrt{2x}\log\sqrt{2x}, \; & \text{if} \; x<\frac{1}{2\e^2}
    \end{cases} \ .
\end{align}
Hence if we set $D_{\rm KL}(\nu_u^\star(\cdot\med\bx)\|\widetilde\nu(\cdot\med\bx)) = f^{-1}(\frac{1}{2}\Delta_{\min})$, we ensure that  $|H(\nu_u^\star(\cdot\med\bX)) - H(\widetilde\nu_u(\cdot\med\bX))| < \frac{1}{2}\Delta_{\min}$.

\underline{\textbf{Finding $f^{-1}(\frac{1}{2}\Delta_{\min})$:}} 

We have the following two cases.
\begin{itemize}
    \item \textbf{$\Delta_{\min} > \frac{2}{\e}(\log |\mcY|+2)$:} In this case, it can be readily verified that
    \begin{align}
        f^{-1}\Big(\frac{1}{2}\Delta_{\min}\Big)\;=\; \frac{1}{2}\left( \frac{\Delta_{\min}}{2(\log|\mcY|+2)}\right)\ .
    \end{align}
    \item \textbf{$\Delta_{\min} \leq \frac{2}{\e}(\log |\mcY|+2)$:} In this case, it can be readily verified that
    \begin{align}
        f^{-1}\Big(\frac{1}{2}\Delta_{\min}\Big)\;=\;\frac{1}{2}\exp\left( 2W_{-1}\left( -\frac{\Delta_{\min}}{2(\log|\mcY|+2)}\right)\right)\ .
    \end{align}
\end{itemize}

\underline{\textbf{Upper bounding $D_{\rm KL}(\nu^\star(\cdot\med\bx)\|\widetilde\nu_u(\cdot\med\bx))$:}}

Next, let us define the normalization constants
\begin{align}
    Z^\star_u(\bx)\;\triangleq\; \sum_{\by}\pi_{\rm ref}(\by\med\bx)\exp\Big( \frac{1}{\beta}r_{\btheta^\star_u}(\bx,\by)\Big)\ ,\;\;\text{and}\;\; \widetilde Z_u(\bx)\;\triangleq\;\sum_{\by}\pi_{\rm ref}(\by\med\bx)\exp\Big( \frac{1}{\beta}r_{\widehat\bB\widehat\bw_u}(\bx,\by)\Big).
\end{align}
With probability at least $1-\delta$, we have
\begin{align}
    D_{\rm KL}&\big(\nu_u^\star(\cdot\med\bx)\|\widetilde\nu_u(\cdot\med\bx) \big)\;\nonumber\\
    &=\;\frac{1}{Z^\star(\bx)}\sum_{\by}\nu^\star_u(\by\med\bx)\log\frac{\exp\Big(\frac{1}{\beta}r_{\btheta^\star_u}(\bx,\by) \Big)\widetilde Z_u(\bx)}{\exp\Big(\frac{1}{\beta}r_{\widehat\bB\widehat\bw_u}(\bx,\by) \Big) Z_u^\star(\bx)}\\
    \label{eq:sample_complexity_4}
    &=\;\frac{1}{Z^\star(\bx)}\sum_{\by}\nu^\star_u(\by\med\bx)\left(\frac{1}{\beta}\big\langle \phi(\bx,\by),\bB^\star\bw_u^\star - \widehat\bB\widehat\bw_u\big\rangle + \log\frac{\widetilde Z_u(\bx)}{Z_u^\star(\bx)}\right)\\
    \label{eq:sample_complexity_5}
    &\leq\;\log\frac{\widetilde Z_u(\bx)}{Z_u^\star(\bx)} + \frac{1}{\beta}\sum_{\by}\nu_u^\star(\by\med\bx)\cdot \| \phi(\bx,\by)\|_{(\Sigma + \lambda\mathbb{I})^{-1}}\cdot\|\widehat\bB\widehat\bw_u -\bB^\star\bw_u^\star\|_{\Sigma + \lambda\mathbb{I}}\\
    \label{eq:sample_complexity_6}
    &\leq\;\log\frac{\widetilde Z_u(\bx)}{Z_u^\star(\bx)} + \frac{1}{\beta}\cdot\eta^{\rm SR}(N,\lambda,\delta)\cdot\E_{\by\sim\nu_u^\star(\by\med\bx)}\Big[\| \phi(\bx,\by)\|_{(\Sigma + \lambda\mathbb{I})^{-1}}\Big]\ ,
\end{align}
where~\eqref{eq:sample_complexity_4} follows from Assumption~\ref{assumption:linear reward}, \eqref{eq:sample_complexity_5} follows from Cauchy-Schwarz, and~\eqref{eq:sample_complexity_6} follows from Lemma~\ref{lemma: est_con_sharedrep}. Furthermore, with probability at least $1-\delta$ we have
\begin{align}
    \log\frac{\widetilde Z_u(\bx)}{Z_u^\star(\bx)}\;&=\;\log\frac{\sum_{\by}\pi_{\rm ref}(\by\med\bx)\exp\Big(\frac{1}{\beta}\big\langle \phi(\bx,\by),\bB^\star\bw_u^\star\big\rangle\Big)\exp\Big(\frac{1}{\beta} \big\langle \phi(\bx,\by),\widehat\bB\widehat\bw_u - \bB^\star\bw_u^\star\big\rangle\Big)}{\sum_{\by}\pi_{\rm ref}\exp\Big(\frac{1}{\beta} \big\langle \phi(\bx,\by),\bB^\star\bw_u^\star\big\rangle\Big)}\\
    &=\;\log\left(\sum_{\by} \nu_u^\star(\by\med\bx)\exp\Big(\frac{1}{\beta} \big\langle \phi(\bx,\by),\widehat\bB\widehat\bw_u - \bB^\star\bw_u^\star\big\rangle\Big)\right)\\
    \label{eq:sample_complexity_7}
    &\leq\;\log\left( \sum_{\by}\nu_u^\star(\by\med\bx)\exp\Big( \frac{1}{\beta}\|\phi(\bx,\by) \|_{(\Sigma + \lambda\mathbb{I})^{-1}}\cdot\|\widehat\bB\widehat\bw_u - \bB^\star\bw_u^\star \|_{\Sigma+\lambda\mathbb{I}}\Big)\right)\\
    \label{eq:sample_complexity_8}
    &\leq\;\log\left( \sum_{\by}\nu_u^\star(\by\med\bx)\exp\Big(\frac{1}{\beta}\eta^{\rm SR}(N,\lambda,\delta)\cdot\| \phi(\bx,\by)\|_{(\Sigma+\lambda\mathbb{I})^{-1}} \Big)\right)\\
    &=\;\log\left(1 + \sum_{\ell\in\N} \frac{1}{\ell !}\left(\frac{\eta(N,\lambda,\delta)}{\beta}\right)^{\ell} \E_{\by\sim\nu_u^\star(\cdot\med\bx)}\Big[ \|\phi(\bx,\by) \|^{\ell}_{(\Sigma + \lambda\mathbb{I})^{-1}}\Big]\right)\\
    \label{eq:sample_complexity_9}
    &\leq\;\sum_{\ell\in\N} \frac{1}{\ell !} \left( \frac{\eta(N,\lambda,\delta)}{\beta}\right)^{\ell} \E_{\by\sim\nu_u^\star(\cdot\med\bx)}\Big[ \|\phi(\bx,\by) \|^{\ell}_{(\Sigma + \lambda\mathbb{I})^{-1}}\Big]\ ,
\end{align}
where~\eqref{eq:sample_complexity_7} follows from Cauchy-Schwarz, \eqref{eq:sample_complexity_8} follows from Lemma~\ref{lemma: est_con_sharedrep}, and~\eqref{eq:sample_complexity_9} follows from the inequality $\log(1+x)\leq x$ for all $x\in\R_+$. Let us set $\lambda=\frac{1}{N}$. According to~\eqref{eq:sample_complexity_9}, with probability at least $1-\delta$, we have
\begin{align}
\label{eq:sample_complexity_10}
    \log\frac{\widetilde Z_u(\bx)}{Z_u^\star(\bx)}\;\leq\;\frac{C_{\rm SR}}{\beta\sqrt{N}}\cdot\sqrt{C_{\delta} + B_{\max}^2}\cdot\E_{\by\sim\nu_u^\star(\cdot\med\bx)}\Big[\| \phi(\bx,\by)\|_{(\Sigma + \lambda\mathbb{I})^{-1}}\Big] + O\Big(\frac{1}{N} \Big)\ .
\end{align}
Furthermore, combining~\eqref{eq:sample_complexity_6} and~\eqref{eq:sample_complexity_10}, we have
\begin{align}
\label{eq:KL_bound}
    D_{\rm KL}\big(\nu_u^\star(\cdot\med\bx)\|\widetilde\nu_u(\cdot\med\bx) \big)\;\leq\; \frac{2}{\beta}\cdot\eta^{\rm SR}\Big(N,\frac{1}{N},\delta \Big)\cdot\E_{\by\sim\nu_u^\star(\cdot\med\bx)}\Big[ \|\phi(\bx,\by) \|_{(\Sigma + \lambda\mathbb{I})^{-1}}\Big] + O\Big( \frac{1}{N}\Big)\ .
\end{align}
Now, following up on case $1$, in order to have $\Big| H\big( \nu^\star_u(\cdot\med\bX)\big) - H\big( \widetilde\nu_u(\cdot\med\bX)\big)\Big| < \frac{\Delta_{\min}}{2}$, leveraging the bound in~\eqref{eq:KL_bound}, we should ensure 
\begin{align}
\label{eq:N_case_1_pre}
    \frac{2C_{\rm SR}}{\beta\sqrt{N}}\cdot\sqrt{C_{\delta} + B_{\max}^2}\cdot\E_{\by\sim\nu_u^\star(\cdot\med\bx)}\Big[ \|\phi(\bx,\by) \|_{(\Sigma + \lambda\mathbb{I})^{-1}}\Big] + O\Big( \frac{1}{N}\Big)\;\leq\;\frac{\Delta_{\min}^2}{8(\log|\mcY|+2)^2}\ .
\end{align}
The condition in~\eqref{eq:N_case_1_pre} can be readily verified to satisfy if, for every subpopulation $u\in[U]$,
\begin{align}
\label{eq:N_case_1}
    N\;\geq\;\;(32C_1)^2\cdot(C_{\delta} + B_{\max}^2)\cdot\frac{(\log|\mcY|+2)^4}{\beta^2\Delta_{\min}^4}\cdot\Big( \max_{\bx\in\mcX}\;\E_{\by\sim\nu_u^\star(\cdot\med\bx)}\Big[ \|\phi(\bx,\by) \|_{(\Sigma + \lambda\mathbb{I})^{-1}}\Big]\Big)^2
\end{align}
for some universal constant $C_1\in\R_+$. Similarly, following up on case $2$, in order to have $\Big| H\big( \nu^\star_u(\cdot\med\bX)\big) - H\big( \widetilde\nu_u(\cdot\med\bX)\big)\Big| < \frac{\Delta_{\min}}{2}$, leveraging the bound in~\eqref{eq:KL_bound}, we should ensure 
\begin{align}
\label{eq:N_case_2_pre}
    \frac{2}{\beta}\cdot\eta^{\rm SR}\Big(N,\frac{1}{N},\delta \Big)\cdot&\E_{\by\sim\nu_u^\star(\cdot\med\bx)}\Big[ \|\phi(\bx,\by) \|_{(\Sigma + \lambda\mathbb{I})^{-1}}\Big] + O\Big( \frac{1}{N}\Big)\;\nonumber\\
    &\leq\;\frac{1}{2}\exp\Big( 2W_{-1}\Big( -\frac{\Delta_{\min}}{2(\log |\mcY| +2)}\Big)\Big)\ .
\end{align}
The condition in~\eqref{eq:N_case_2_pre} can be readily verified to satisfy if, for every $u\in[U]$,
\begin{align}
\label{eq:N_case_2}
    N\;\geq\;\left(\frac{8C_2}{\beta}\right)^2\cdot(C_\delta + B_{\max})^2 &\exp\left( -4W_{-1}\left( -\frac{\Delta_{\min}}{2(\log|\mcY|+2)}\right)\right)\nonumber\\
    &\times\Big(\max_{\bx\in\mcX}\; \E_{\by\sim\nu_u^\star(\cdot\med\bx)}\Big[ \|\phi(\bx,\by) \|_{(\Sigma + \lambda\mathbb{I})^{-1}}\Big]\Big)^2\ ,
\end{align}
where $C_2\in\R_+$ is a universal constant. Let us set $N_{\rm MaxMin}$ as defined in Theorem~\ref{theorem: sample complexity}. For $N>N_{\rm MaxMin}$ samples, we have
\begin{align}
\label{eq:sample_complexity_11}
    \Big| H\big( \nu^\star_u(\cdot\med\bX)\big) - H\big( \widetilde\nu_u(\cdot\med\bX)\big)\Big| < \frac{\Delta_{\min}}{2}\ .
\end{align}
Hence, we have
\begin{align}
    \P(\widehat u\neq u^\star)\;&\leq\; \sum_{u\neq u^\star} \P(\widehat u = u)\\
    \label{eq:sample_complexity_12}
    &= \sum_{u\neq u^\star} \P\Big( H\big(\widetilde\nu_u(\cdot\med\bX)\big) > H\big(\widetilde\nu_v(\cdot\med\bX) \big)\;\;\forall\;v\neq u\Big)\\
    &\leq\;\sum_{u\neq u^\star}\P\Big(H\big(\widetilde\nu_u(\cdot\med\bX) > H\big( \widetilde\nu_{u^\star}(\cdot\med\bX)\big) \big) \Big)\\
    &\stackrel{\eqref{eq:sample_complexity_11}}{\leq}\sum_{u\neq u^\star} \P\Big( H\big(\nu^\star_u(\cdot\med\bX)\big) + \frac{1}{2}\Delta_{\min} > H\big( \nu^\star_{u^\star}(\cdot\med\bX)\big) - \frac{1}{2}\Delta_{\min}\Big)\\
    \label{eq:sample_complexity_13}
    &=\;0\ ,
\end{align}
where~\eqref{eq:sample_complexity_12} follows from Lemma~\ref{lemma:equivalence_of_u} and~\eqref{eq:sample_complexity_13} holds since $\Delta_{\min}$ is the minimum gap in the conditional entropies between Gibbs distributions across subpopulations. The proof is completed by combining~\eqref{eq:A_3} and~\eqref{eq:sample_complexity_12}.
\section{Experimental Details}
\label{appendix: experiments}

\begin{table*}[t]
\small
\centering
\setlength{\tabcolsep}{4pt}
\renewcommand{\arraystretch}{1.1}
\begin{tabular}{ccccccc}
\toprule
\multirow{2}{*}{\shortstack{Minority\\Proportion}} 
& \multicolumn{3}{c}{Mean Minority Score}
& \multicolumn{3}{c}{Mean Majority Score}\\
\cmidrule(lr){2-4} \cmidrule(lr){5-7}
& MaxMin & SharedRep & Gold
& MaxMin & SharedRep & Gold\\
\midrule
0.01 & \num{0.589612} $\pm$ \num{0.000616} & \textbf{\num{0.687198}} $\pm$ \num{0.008603} & \multirow{4}{*}{\num{0.8084136299465374} $\pm$ \num{0.001445409734884519}} & \num{0.445243} $\pm$ \num{0.000768} & \textbf{\num{0.693248}} $\pm$ \num{0.000373} & \multirow{4}{*}{\num{0.847841396429763} $\pm$ \num{0.0005405260297135759}}\\
0.05 & \num{0.628440} $\pm$ \num{0.000601} & \textbf{\num{0.689723}} $\pm$ \num{0.014984} &  & \num{0.566553} $\pm$ \num{0.000732} & \textbf{\num{0.670058}} $\pm$ \num{0.000373} & \\
0.10 & \num{0.647401} $\pm$ \num{0.000590} & \textbf{\num{0.706462}} $\pm$ \num{0.012821} &  & \num{0.591683} $\pm$ \num{0.000746} & \textbf{\num{0.766877}} $\pm$ \num{0.000332} & \\
0.20 & \num{0.673009} $\pm$ \num{0.000572} & \textbf{\num{0.724814}} $\pm$ \num{0.012303} &  & \num{0.523052} $\pm$ \num{0.000664} & \textbf{\num{0.750169}} $\pm$ \num{0.000305} & \\
\bottomrule
\end{tabular}
\caption{
Comparison of mean scores between MaxMin- and SharedRep-RLHF ($K=2$) across different minority proportions for the IMDb dataset~\cite{maas2011learning}. Mean scores are shown separately for minority and majority groups.
    }
\label{tab:mean_score_IMDb_extended}
\end{table*}

\begin{table*}[t]
\small
\centering
\setlength{\tabcolsep}{4pt}
\renewcommand{\arraystretch}{1.1}
\begin{tabular}{ccccc}
\toprule
\multirow{2}{*}{\shortstack{Minority\\Proportion}} 
& \multicolumn{2}{c}{Minority Win Rate (\%)}
& \multicolumn{2}{c}{Majority Win Rate (\%)} \\
\cmidrule(lr){2-3} \cmidrule(lr){4-5}
& MaxMin & SharedRep
& MaxMin & SharedRep \\
\midrule
0.01 & 21.38 & \textbf{25.26} & \textbf{26.67} & 21.31 \\
0.05 & 25.21 & \textbf{26.01} & \textbf{34.25} & 15.37 \\
0.10 & 28.31 & \textbf{31.80} & \textbf{40.14} & 31.95 \\
0.20 & 23.97 & \textbf{31.49} & 16.52 & \textbf{26.04} \\
\bottomrule
\end{tabular}
\caption{
Win rates (\%) between MaxMin- and SharedRep-RLHF ($K=2$) across different minority proportions for the IMDb dataset~\cite{maas2011learning}. Win rates indicate the fraction of instances in which the algorithm responses are (score-wise) better than the gold reward skyline.
}
\label{tab:winrate_IMDb_extended}
\end{table*}

\begin{figure}[t]
    \centering
    \pgfplotsset{compat=newest}

\begin{tikzpicture}[yshift=-1.5cm]
  \begin{groupplot}[
    group style={
      group size=2 by 2,
      horizontal sep=2.6cm,
      vertical sep=2.6cm,
    },
    xlabel={$K$},
    ylabel={group score},
    symbolic x coords={2,4,16,32},
    xtick=data,
    ymajorgrids=true,
    xmajorgrids=true,
    grid style=dotted,
    height=4cm,
    width=5cm,
    enlarge x limits=0.15,
    error bars/y dir=both,
    error bars/y explicit,
  ]

  \nextgroupplot[
    title={Minority proportion$=5$\%},
    legend to name=legendtop,
    legend style={
      nodes={scale=0.9, transform shape},
      at={(0.5,-0.25)}, anchor=north,
      legend columns=2,
      /tikz/every even column/.append style={column sep=0.6cm}
    },
  ]
  \addplot[
    black,
    line width=1pt,
    mark=triangle*,
    error bars/.cd,
    y dir=both,
    y explicit
  ] table [
    x=k,
    y=mean_majority_score,
    y error expr=\thisrowno{4},
    col sep=comma
  ]{figures/ablation/imdb/data/score_prop_0.05.csv};
  \addlegendentry{Majority group score}

  \addplot[
    black,
    dotted,
    line width=1pt,
    mark=square*,
    error bars/.cd,
    y dir=both,
    y explicit
  ] table [
    x=k,
    y=mean_minority_score,
    y error expr=\thisrowno{2},
    col sep=comma
  ]{figures/ablation/imdb/data/score_prop_0.05.csv};
  \addlegendentry{Minority group score}

  \nextgroupplot[
    title={Minority proportion$=10$\%},
  ]
  \addplot[
    black,
    line width=1pt,
    mark=triangle*,
    error bars/.cd,
    y dir=both,
    y explicit
  ] table [
    x=k,
    y=mean_majority_score,
    y error expr=\thisrowno{4},
    col sep=comma
  ]{figures/ablation/imdb/data/score_prop_0.1.csv};

  \addplot[
    black,
    dotted,
    line width=1pt,
    mark=square*,
    error bars/.cd,
    y dir=both,
    y explicit
  ] table [
    x=k,
    y=mean_minority_score,
    y error expr=\thisrowno{2},
    col sep=comma
  ]{figures/ablation/imdb/data/score_prop_0.1.csv};

    \nextgroupplot[
      ylabel={win rate},
      ymin=0.0,
      ymax=0.6,
      legend to name=legendbottom, 
      legend style={
        nodes={scale=0.9, transform shape},
        legend columns=2,
        /tikz/every even column/.append style={column sep=0.6cm}
      },
    ]
    \addplot+[
      ybar,
      bar width=10pt,
      bar shift=-5pt,
      draw=black,
      mark=none,
      pattern=north east lines,
      pattern color=black,
      legend image code/.code={
        \draw[draw=black,pattern=north east lines,pattern color=black] 
          (0cm,-0.1cm) rectangle (0.3cm,0.1cm);
      }
    ] table[x=k, y=maj_win, col sep=comma] {figures/ablation/imdb/data/winrate_prop_0.05.csv};
    \addlegendentry{Majority win rate}
    
    \addplot+[
      ybar,
      bar width=10pt,
      bar shift=5pt,
      draw=black,
      mark=none,
      pattern=crosshatch,
      pattern color=black,
      legend image code/.code={
        \draw[draw=black,pattern=crosshatch,pattern color=black] 
          (0cm,-0.1cm) rectangle (0.3cm,0.1cm);
      }
    ] table[x=k, y=min_win, col sep=comma] {figures/ablation/imdb/data/winrate_prop_0.05.csv};
    \addlegendentry{Minority win rate}

    \nextgroupplot[
      ylabel={win rate},
      ymin=0.0,
      ymax=0.6
    ]
    \addplot+[
      ybar,
      bar width=10pt,
      bar shift=-5pt,
      mark=none,
      draw=black,
      pattern=north east lines,
      pattern color=black
    ] table[x=k, y=maj_win, col sep=comma] {figures/ablation/imdb/data/winrate_prop_0.1.csv};
    
    \addplot+[
      ybar,
      bar width=10pt,
      bar shift=5pt,
      mark=none,
      draw=black,
      pattern=crosshatch,
      pattern color=black
    ] table[x=k, y=min_win, col sep=comma] {figures/ablation/imdb/data/winrate_prop_0.1.csv};

  \end{groupplot}

  \node at ($(group c1r1)!.5!(group c2r1)+(0,2.3cm)$) {\ref{legendtop}};

  \node at ($(group c1r2)!.5!(group c2r2)+(0,2.2cm)$) {\ref{legendbottom}};

\end{tikzpicture}
    \caption{Ablation in $K$ computed for the IMDb dataset~\cite{maas2011learning}. \textbf{Top:} Average scores achieved by SharedRep-RLHF for various values of the inner dimension $K$, when minority proportion sizes are set to $1$\% (on the left), and $20$\% (on the right). \textbf{Bottom:} Win rates (fractions) achieved by SharedRep-RLHF for various values of the inner dimension $K$, when minority proportion sizes are set to $1$\% (on the left), and $20$\% (on the right).}
    \label{fig:ablation}
\end{figure}

\begin{table}[t]
\small
\centering
\setlength{\tabcolsep}{4pt}
\begin{tabular}{lll}
\toprule
\textbf{Hyperparameter} & \textbf{Reward} & \textbf{PPO} \\
\midrule
Optimizer & AdamW ($\beta_1$=0.9, $\beta_2$=0.999, $\varepsilon$=1e-8) & AdamW ($\beta_1$=0.9, $\beta_2$=0.999, $\varepsilon$=1e-8) \\
Learning Rate & $5\times 10^{-3}$ & $3\times10^{-6}$ \\
LR Scheduler & Cosine Annealing (default) & Cosine Annealing (default) \\
Epochs & $5$ & $3$ \\
Batch Size & $512$ & $256$ \\
KL regularization ($\beta$) & - & $0.05$\\
\bottomrule
\end{tabular}
\caption{IMDb training hyperparameters.}
\label{tab:imdb_hyperparams}
\end{table}

\subsection{Controlled Sentiment Analysis}
\paragraph{Setting.} We use the IMDb dataset~\cite{maas2011learning}, a large corpus of movie reviews labeled as either positive or negative. Each review consists of natural language text expressing subjective opinions. Our goal is to train models that can generate review continuations that reflect different group preferences, specifically, for sentiment and brevity. To implement this, we extract prompts by truncating each review in the training split to its first $2-8$ tokens, following the method advocated by~\citeauthor{rafailov2023direct}. These truncated snippets serve as inputs to the language model. To simulate pluralistic preferences, we randomly partition the dataset into two synthetic sub-populations: a {\em majority} and a {\em minority}. The majority group values only {\em brevity} (interchangeably, conciseness), while the minority group values both {\em brevity} and {\em positive sentiment}. For reward modeling, we assign each generated response a group-specific gold score. The majority score is computed as a normalized measure of conciseness based on character length. The minority score is a weighted combination of sentiment and conciseness, with a $70$\% weight on the sentiment score (obtained via the pretrained \texttt{lvwerra/distilbert-imdb} classifier) and $30$\% on conciseness. Preference labels are then derived by comparing gold scores across response pairs: the response with the higher score is labeled as ``chosen''. For the RL fine-tuning step, we use PPO~\cite{schulman2017proximal} to learn the optimal policy. A complete list of hyperparameters can be found in Table~\ref{tab:imdb_hyperparams}.

\paragraph{Results.} Tables~\ref{tab:mean_score_IMDb_extended} and~\ref{tab:winrate_IMDb_extended} extend Table~\ref{tab:mean_score_winrate_IMDb}, presenting average scores and win rates for both majority and minority sub-populations across varying levels of minority representation. While our primary objective is to safeguard the interests of all annotators by improving the {\em average minority score}, we find that SharedRep-RLHF not only achieves this goal but also preserves — and in some cases enhances — majority group performance. Specifically, SharedRep-RLHF improves the average majority score at all minority proportions, and notably increases the win rate for the majority group when the minority constitutes $20$\% of the population. While MaxMin-RLHF achieves a better win rate for the {\em majority group} in several instances, note that this is not a drawback of SharedRep-RLHF -- having better win rates is possibly indicative of a biased allegiance towards the preferences of the majority, at the cost of minority's preferences. We further perform an ablation study on the inner dimension hyperparameter $K$ of the SharedRep-RLHF algorithm, with results summarized in Figure~\ref{fig:ablation}. We observe that $K=32$ yields the largest gain in average minority score at $1$\% as well as $20$\% levels. Overall, $K=2$ strikes a favorable balance across settings in terms of both the average minority as well as majority scores, and achieves the largest win rate for both the majority and minority subpopulations. Hence, $K=2$ is used for all reported results in Section~\ref{sec:experiments}.

\begin{table*}[t]
\centering
\small
\setlength{\tabcolsep}{4pt}
\begin{tabular}{ccccccc}
\toprule
\multirow{2}{*}{\shortstack{Minority\\Proportion}} 
& \multicolumn{3}{c}{Mean Minority Score}
& \multicolumn{3}{c}{Mean Majority Score}\\
\cmidrule(lr){2-4} \cmidrule(lr){5-7} 
& MaxMin & SharedRep & Gold
& MaxMin & SharedRep & Gold \\
\midrule
0.01 & \num{0.239025} $\pm$ \num{0.002723} & \textbf{\num{0.310037}} $\pm$ \num{0.002126} & \multirow{4}{*}{\num{0.41453692978445056} $\pm$ \num{0.0032619289734337528}} &\num{0.405416} $\pm$ \num{0.003150} & \textbf{\num{0.597923}} $\pm$ \num{0.002213} & \multirow{4}{*}{\num{0.7843923115308477} $\pm$ \num{0.002034416314563246}} \\
0.05 & \num{0.136607} $\pm$ \num{0.001588} & \textbf{\num{0.287745}} $\pm$ \num{0.002123} & &\textbf{\num{0.687061}} $\pm$ \num{0.001856} & \num{0.608934} $\pm$ \num{0.002378} &  \\
0.10 & \num{0.194189} $\pm$ \num{0.001730} & \textbf{\num{0.293033}} $\pm$ \num{0.002352} & &\num{0.574142} $\pm$ \num{0.002093	} & \textbf{\num{0.587270}} $\pm$ \num{0.002454} & \\
0.15 & \textbf{\num{0.316723}} $\pm$ \num{0.002699}  & \num{0.277359} $\pm$ \num{0.002421}  & &\textbf{\num{0.538222}} $\pm$ \num{0.003868} & \num{0.529822} $\pm$ \num{0.003575} & \\
0.20 & \textbf{\num{0.296033}} $\pm$ \num{0.003889} & \num{0.269524} $\pm$ \num{0.002006} & &\num{0.522338} $\pm$ \num{0.002223} & \textbf{\num{0.599178	}} $\pm$ \num{0.002232} & \\
\bottomrule
\end{tabular}
\caption{
Comparison of mean scores between MaxMin- and SharedRep-RLHF ($K=16$) across different minority proportions for GSM8K~\cite{cobbe2021training}. Mean scores are shown separately for minority and majority groups.
}
\label{tab:mean_score_winrate_gsm8k_extended}
\end{table*}

\begin{table*}[t]
\small
\centering
\setlength{\tabcolsep}{4pt}
\centering
\begin{tabular}{ccccc}
\toprule
\multirow{2}{*}{\shortstack{Minority\\Proportion}} 
& \multicolumn{2}{c}{Minority Win Rate (\%)}
& \multicolumn{2}{c}{Majority Win Rate (\%)} \\
\cmidrule(lr){2-3} \cmidrule(lr){4-5}
& MaxMin & SharedRep & MaxMin & SharedRep \\
\midrule
0.01 & 19.92 & \textbf{29.72} & 1.91 & \textbf{7.25} \\
0.05 & 4.93 & \textbf{25.32} & \textbf{30.42} & 9.5 \\
0.10 & 11.43 & \textbf{28.90} & 4.69	 & \textbf{8.15} \\
0.15 & \textbf{30.52} & 24.92 & \textbf{10.42}	 & 7.86 \\
0.20 & \textbf{31.65} & 22.76 & 3.18 & \textbf{8.2} \\
\bottomrule
\end{tabular}
\caption{
Win rates (\%) between MaxMin- and SharedRep-RLHF ($K=16$) across different minority proportions for GSM8K~\cite{cobbe2021training}. Win rates indicate the fraction of instances in which the algorithm responses are (score-wise) better than the gold reward skyline.
}
\label{tab:winrate_gsm8k_extended}
\end{table*}

\begin{figure}[htb]
    \centering
    \pgfplotsset{compat=newest}

\begin{tikzpicture}[yshift=-1.5cm]
  \begin{groupplot}[
    group style={
      group size=2 by 2,
      horizontal sep=2.6cm,
      vertical sep=2.6cm,
    },
    xlabel={$K$},
    ylabel={group score},
    symbolic x coords={2,8,16},
    xtick=data,
    ymajorgrids=true,
    xmajorgrids=true,
    grid style=dotted,
    height=4cm,
    width=5cm,
    enlarge x limits=0.15,
    error bars/y dir=both,
    error bars/y explicit,
  ]

  \nextgroupplot[
    title={Minority proportion$=5$\%},
    legend to name=legendtop,
    legend style={
      nodes={scale=0.9, transform shape},
      at={(0.5,-0.25)}, anchor=north,
      legend columns=2,
      /tikz/every even column/.append style={column sep=0.6cm}
    },
  ]
  \addplot[
    black,
    line width=1pt,
    mark=triangle*,
    error bars/.cd,
    y dir=both,
    y explicit
  ] table [
    x=k,
    y=mean_majority_score,
    y error expr=\thisrowno{4},
    col sep=comma
  ]{figures/ablation/gsm8k/data/score_prop_0.05.csv};
  \addlegendentry{Majority group score}

  \addplot[
    black,
    dotted,
    line width=1pt,
    mark=square*,
    error bars/.cd,
    y dir=both,
    y explicit
  ] table [
    x=k,
    y=mean_minority_score,
    y error expr=\thisrowno{2},
    col sep=comma
  ]{figures/ablation/gsm8k/data/score_prop_0.05.csv};
  \addlegendentry{Minority group score}

  \nextgroupplot[
    title={Minority proportion$=10$\%},
  ]
  \addplot[
    black,
    line width=1pt,
    mark=triangle*,
    error bars/.cd,
    y dir=both,
    y explicit
  ] table [
    x=k,
    y=mean_majority_score,
    y error expr=\thisrowno{4},
    col sep=comma
  ]{figures/ablation/gsm8k/data/score_prop_0.1.csv};

  \addplot[
    black,
    dotted,
    line width=1pt,
    mark=square*,
    error bars/.cd,
    y dir=both,
    y explicit
  ] table [
    x=k,
    y=mean_minority_score,
    y error expr=\thisrowno{2},
    col sep=comma
  ]{figures/ablation/gsm8k/data/score_prop_0.1.csv};

    \nextgroupplot[
      ylabel={win rate},
      ymin=0.0,
      ymax=0.4,
      legend to name=legendbottom, 
      legend style={
        nodes={scale=0.9, transform shape},
        legend columns=2,
        /tikz/every even column/.append style={column sep=0.6cm}
      },
    ]
    \addplot+[
      ybar,
      bar width=10pt,
      bar shift=-5pt,
      draw=black,
      mark=none,
      pattern=north east lines,
      pattern color=black,
      legend image code/.code={
        \draw[draw=black,pattern=north east lines,pattern color=black] 
          (0cm,-0.1cm) rectangle (0.3cm,0.1cm);
      }
    ] table[x=k, y=maj_win, col sep=comma] {figures/ablation/gsm8k/data/winrate_prop_0.05.csv};
    \addlegendentry{Majority win rate}
    
    \addplot+[
      ybar,
      bar width=10pt,
      bar shift=5pt,
      draw=black,
      mark=none,
      pattern=crosshatch,
      pattern color=black,
      legend image code/.code={
        \draw[draw=black,pattern=crosshatch,pattern color=black] 
          (0cm,-0.1cm) rectangle (0.3cm,0.1cm);
      }
    ] table[x=k, y=min_win, col sep=comma] {figures/ablation/gsm8k/data/winrate_prop_0.05.csv};
    \addlegendentry{Minority win rate}

    \nextgroupplot[
      ylabel={win rate},
      ymin=0.0,
      ymax=0.4
    ]
    \addplot+[
      ybar,
      bar width=10pt,
      bar shift=-5pt,
      mark=none,
      draw=black,
      pattern=north east lines,
      pattern color=black
    ] table[x=k, y=maj_win, col sep=comma] {figures/ablation/gsm8k/data/winrate_prop_0.1.csv};
    
    \addplot+[
      ybar,
      bar width=10pt,
      bar shift=5pt,
      mark=none,
      draw=black,
      pattern=crosshatch,
      pattern color=black
    ] table[x=k, y=min_win, col sep=comma] {figures/ablation/gsm8k/data/winrate_prop_0.1.csv};

  \end{groupplot}

  \node at ($(group c1r1)!.5!(group c2r1)+(0,2.3cm)$) {\ref{legendtop}};

  \node at ($(group c1r2)!.5!(group c2r2)+(0,2.2cm)$) {\ref{legendbottom}};

\end{tikzpicture}
    \caption{Ablation in $K$ computed for the GSM8K dataset~\cite{cobbe2021training}. \textbf{Top:} Average scores achieved by SharedRep-RLHF for various values of the inner dimension $K$, when minority proportion sizes are set to $1$\% (on the left), and $20$\% (on the right). \textbf{Bottom:} Win rates (fractions) achieved by SharedRep-RLHF for various values of the inner dimension $K$, when minority proportion sizes are set to $1$\% (on the left), and $20$\% (on the right).}
    \label{fig:ablation_gsm8k}
\end{figure}

\begin{table}[t]
\small
\centering
\setlength{\tabcolsep}{4pt}
\begin{tabular}{lll}
\toprule
\textbf{Hyperparameter} & \textbf{Reward} & \textbf{GRPO} \\
\midrule
Optimizer & AdamW ($\beta_1$=0.9, $\beta_2$=0.999, $\varepsilon$=1e-8) & AdamW ($\beta_1$=0.9, $\beta_2$=0.999, $\varepsilon$=1e-8) \\
Learning Rate & $5\times 10^{-4}$ & $1\times10^{-5}$ \\
LR Scheduler & Linear & Linear \\
Epochs & $3$ & $2$ \\
Batch Size & $64$ & $32$ \\
KL regularization ($\beta$) & - & $0.05$\\
\bottomrule
\end{tabular}
\caption{GSM8K and HH training hyperparameters.}
\label{tab:gsm8k_hh_hyperparams}
\end{table}

\subsection{Mathematical Reasoning}
\paragraph{Setting.} We use the GSM8K dataset~\cite{cobbe2021training}, a collection of high-quality grade-school math problems that require multi-step arithmetic reasoning. Each example consists of a natural language question followed by a detailed CoT solution and a final numeric answer. Our goal is to train models to produce completions that align with different stylistic preferences -- either concise or elaborate Socratic reasoning -- while maintaining correctness. To implement this, we prepend each question with a fixed two-shot CoT prefix to encourage the model to generate step-by-step reasoning. We simulate pluralistic preferences by defining two sub-populations: the {\em majority}, which favors correctness and brevity, and the {\em minority}, which favors correctness and Socratic-style elaboration. Each response is evaluated on three dimensions: accuracy, length, and Socratic quality. Accuracy is determined by extracting and matching the final numeric answer with the ground truth. Conciseness is measured using normalized response length. Socratic quality is scored using a reward model fine-tuned to detect elaborative, step-by-step reasoning. Specifically, we train a GPT2-large model on the ``Socratic'' split of GSM8K until convergence. We then use this model as the gold reward for evaluating the Socratic score. We compute group-specific gold scores using weighted combinations of these dimensions. The majority score assigns $20$\% weight to correctness and $80$\% to brevity. The minority score assigns $20$\% weight to correctness and $80$\% to Socratic quality. As with the sentiment task, preference labels are created by comparing group-specific gold scores across response pairs. We use GRPO~\cite{shao2024deepseekmath} for the RL fine-tuning step to learn an optimal policy post reward-modeling. A list of the hyperparameters used for this experiment can be found in Table~\ref{tab:gsm8k_hh_hyperparams}.

\paragraph{Results.} Tables~\ref{tab:mean_score_winrate_gsm8k_extended} and~\ref{tab:winrate_gsm8k_extended}, extended versions of Table~\ref{tab:mean_score_winrate_GSM8K}, report average scores and win rates for both majority and minority sub-populations on the GSM8K dataset, across varying levels of minority representation. Our central objective is to elevate the {\em average minority score}, while maintaining competitive performance for the majority group. We find that SharedRep-RLHF substantially improves minority outcomes at low minority proportions (e.g., $1$\%, $5$\%, and $10$\%), outperforming MaxMin-RLHF by wide margins in both score and win rate. In particular, SharedRep-RLHF increases the minority win rate from $4.93$\% to $25.32$\% at $5$\% minority proportion. At higher proportions ($15$\%, $20$\%), MaxMin-RLHF achieves slightly higher minority scores, but this comes at the cost of reduced majority performance. In contrast, SharedRep-RLHF maintains or improves majority scores across all settings and notably increases the majority win rate at $20$\% minority representation (from $3.18$\% to $8.20$\%). While MaxMin-RLHF occasionally outperforms SharedRep-RLHF on majority win rate at lower minority proportions, this is arguably due to its skewed optimization toward majority preferences. Overall, SharedRep-RLHF better balances competing group interests, especially under imbalanced preference distributions. Additionally, we present an ablation study of the inner dimension $K$ of the SharedRep-RLHF algorithm on GSM8K, shown in Figure~\ref{fig:ablation_gsm8k}. The results capture performance variations along two axes: average score and win rate. At a $5$\% minority proportion, model performance remains relatively stable across different values of $K$. At a $10$\% minority proportion, $K=8$ achieves the highest performance on both metrics. Notably, performance at $K=8$ and $K=16$ is comparable for both proportion levels, suggesting robustness to moderate changes in the inner dimension.

\begin{figure}[htb]
    \centering
    \pgfplotsset{compat=newest}

\begin{tikzpicture}[yshift=-1.5cm]
  \begin{groupplot}[
    group style={
      group size=2 by 2,
      horizontal sep=2.6cm,
      vertical sep=2.6cm,
    },
    xlabel={$K$},
    ylabel={group score},
    symbolic x coords={2,8,16},
    xtick=data,
    ymajorgrids=true,
    xmajorgrids=true,
    grid style=dotted,
    height=4cm,
    width=5cm,
    enlarge x limits=0.15,
    error bars/y dir=both,
    error bars/y explicit,
  ]

  \nextgroupplot[
    title={Minority proportion$=1$\%},
    legend to name=legendtop,
    legend style={
      nodes={scale=0.9, transform shape},
      at={(0.5,-0.25)}, anchor=north,
      legend columns=2,
      /tikz/every even column/.append style={column sep=0.6cm}
    },
  ]
  \addplot[
    black,
    line width=1pt,
    mark=triangle*,
    error bars/.cd,
    y dir=both,
    y explicit
  ] table [
    x=k,
    y=mean_majority_score,
    y error expr=\thisrowno{4},
    col sep=comma
  ]{figures/ablation/hh/data/score_prop_0.01.csv};
  \addlegendentry{Majority group score}

  \addplot[
    black,
    dotted,
    line width=1pt,
    mark=square*,
    error bars/.cd,
    y dir=both,
    y explicit
  ] table [
    x=k,
    y=mean_minority_score,
    y error expr=\thisrowno{2},
    col sep=comma
  ]{figures/ablation/hh/data/score_prop_0.01.csv};
  \addlegendentry{Minority group score}

  \nextgroupplot[
    title={Minority proportion$=10$\%},
  ]
  \addplot[
    black,
    line width=1pt,
    mark=triangle*,
    error bars/.cd,
    y dir=both,
    y explicit
  ] table [
    x=k,
    y=mean_majority_score,
    y error expr=\thisrowno{4},
    col sep=comma
  ]{figures/ablation/hh/data/score_prop_0.1.csv};

  \addplot[
    black,
    dotted,
    line width=1pt,
    mark=square*,
    error bars/.cd,
    y dir=both,
    y explicit
  ] table [
    x=k,
    y=mean_minority_score,
    y error expr=\thisrowno{2},
    col sep=comma
  ]{figures/ablation/hh/data/score_prop_0.1.csv};

    \nextgroupplot[
      ylabel={win rate},
      ymin=0.0,
      legend to name=legendbottom, 
      legend style={
        nodes={scale=0.9, transform shape},
        legend columns=2,
        /tikz/every even column/.append style={column sep=0.6cm}
      },
    ]
    \addplot+[
      ybar,
      bar width=10pt,
      bar shift=-5pt,
      draw=black,
      mark=none,
      pattern=north east lines,
      pattern color=black,
      legend image code/.code={
        \draw[draw=black,pattern=north east lines,pattern color=black] 
          (0cm,-0.1cm) rectangle (0.3cm,0.1cm);
      }
    ] table[x=k, y=maj_win, col sep=comma] {figures/ablation/hh/data/winrate_prop_0.01.csv};
    \addlegendentry{Majority win rate}
    
    \addplot+[
      ybar,
      bar width=10pt,
      bar shift=5pt,
      draw=black,
      mark=none,
      pattern=crosshatch,
      pattern color=black,
      legend image code/.code={
        \draw[draw=black,pattern=crosshatch,pattern color=black] 
          (0cm,-0.1cm) rectangle (0.3cm,0.1cm);
      }
    ] table[x=k, y=min_win, col sep=comma] {figures/ablation/hh/data/winrate_prop_0.01.csv};
    \addlegendentry{Minority win rate}

    \nextgroupplot[
      ylabel={win rate},
      ymin=0.0,
    ]
    \addplot+[
      ybar,
      bar width=10pt,
      bar shift=-5pt,
      mark=none,
      draw=black,
      pattern=north east lines,
      pattern color=black
    ] table[x=k, y=maj_win, col sep=comma] {figures/ablation/hh/data/winrate_prop_0.1.csv};
    
    \addplot+[
      ybar,
      bar width=10pt,
      bar shift=5pt,
      mark=none,
      draw=black,
      pattern=crosshatch,
      pattern color=black
    ] table[x=k, y=min_win, col sep=comma] {figures/ablation/hh/data/winrate_prop_0.1.csv};

  \end{groupplot}

  \node at ($(group c1r1)!.5!(group c2r1)+(0,2.3cm)$) {\ref{legendtop}};

  \node at ($(group c1r2)!.5!(group c2r2)+(0,2.2cm)$) {\ref{legendbottom}};

\end{tikzpicture}
    \caption{Ablation in $K$ computed for the HH dataset~\cite{askell2021general}. \textbf{Top:} Average scores achieved by SharedRep-RLHF for two values of the inner dimension $K$, when minority proportion sizes are set to $1$\% (on the left), and $10$\% (on the right). \textbf{Bottom:} Win rates (fractions) achieved by SharedRep-RLHF for various values of the inner dimension $K$, when minority proportion sizes are set to $1$\% (on the left), and $10$\% (on the right).
    }
    \label{fig:ablation_hh}
\end{figure}

\begin{table*}[t]
\small
\setlength{\tabcolsep}{4pt}
\centering
\begin{tabular}{ccccccccccc}
\toprule
\multirow{2}{*}{\shortstack{Minority\\Proportion}} 
& \multicolumn{3}{c}{Mean Minority Score}
& \multicolumn{3}{c}{Mean Majority Score}\\
\cmidrule(lr){2-4} \cmidrule(lr){5-7}
& MaxMin & SharedRep & Gold
& MaxMin & SharedRep & Gold\\
\midrule
0.01 & \num{-0.038328} $\pm$ \num{0.024078} & \textbf{\num{0.017299}} $\pm$ \num{0.023905} & \multirow{2}{*}{\num{0.03199258257120318} $\pm$ \num{0.03844603036476687}} & \textbf{\num{-0.119327}} $\pm$ \num{0.013543} & \textbf{\num{-0.328709}} $\pm$ \num{0.015287} & \multirow{2}{*}{\num{0.44220055792046115} $\pm$ \num{0.018785883755753047}}\\
0.10 & \num{-0.056569} $\pm$ \num{0.024481} & \textbf{\num{-0.029671}} $\pm$ \num{0.024076} &  & \textbf{\num{-0.221450}} $\pm$ \num{0.014718} & \num{-0.243773} $\pm$ \num{0.014965} \\
\bottomrule
\end{tabular}
\caption{Comparison of mean scores between MaxMin- and SharedRep-RLHF ($K=2$) for two minority proportions in HH~\cite{askell2021general}. Mean scores are shown separately for minority and majority groups.}
\label{tab:HH_scores}
\end{table*}

\begin{table*}[t]
\small
\setlength{\tabcolsep}{4pt}
\centering
\begin{tabular}{ccccc}
\toprule
\multirow{2}{*}{\shortstack{Minority\\Proportion}} 
& \multicolumn{2}{c}{Minority Win Rate (\%)}
& \multicolumn{2}{c}{Majority Win Rate (\%)} \\
\cmidrule(lr){2-3} \cmidrule(lr){4-5}
& MaxMin & SharedRep & MaxMin & SharedRep \\
\midrule
0.01 & 33 & \textbf{34} & 41 & \textbf{43} \\
0.10 &  30 & \textbf{32} & \textbf{46} & $44$ \\
\bottomrule
\end{tabular}
\caption{Win rates (\%) between MaxMin- and SharedRep-RLHF ($K=2$) for two minority proportions in HH~\cite{askell2021general}. Win rates (rounded to two significant digits) indicate the fraction of instances in which the algorithm responses are (score-wise) better than the gold reward skyline.}
\label{tab: HH_winrates}
\end{table*}

\subsection{Single-Turn Dialogue}
\paragraph{Setting.} We perform preliminary experiments with the Anthropic-HH dataset~\cite{askell2021general}, a benchmark for human preference modeling in single-turn assistant interactions. Each example consists of a pair of dialogues between a human and an assistant marked as ``chosen'' and ``rejected'', which differ only in the ultimate assistant response. To simulate diverse preferences, we randomly subsample $10,000$ rows from the ``train'' split of the dataset. For each row, we extract the prompt, defined as the text leading up to the final assistant turn, and ask the model to generate the assistant response. We use \texttt{TinyLlama/TinyLlama-1.1B-Chat-v1.0} as the policy model, which is first fine-tuned on the chosen split of the dataset, and a preference dataset is created by generating a pair of responses from the SFT model. As reward models, we use GPT2-large with a frozen backbone. Similarly to the sentiment analysis and mathematical reasoning tasks, for MaxMin-RLHF, we attach a regression head to the frozen backbone per group, whereas for SharedRep-RLHF, we attach a linear layer followed by a pair of regression heads to the frozen backbone for reward-modeling. We define two sub-populations with distinct alignment goals: the {\em majority}, which values responses that are both helpful and harmless ($30$\% helpfulness, $70$\% harmlessness), and the {\em minority}, which values helpfulness exclusively. To operationalize these preferences, we use two pretrained reward models: \texttt{Ray2333/gpt2-large-helpful-reward-model} for helpfulness, and \texttt{Ray2333/gpt2-large-harmless-reward-model} for harmlessness. Each response is scored by both reward models, and group-specific gold scores are computed as weighted combinations of the two. As with the mathematical reasoning task, preference labels are assigned by comparing group-specific gold scores between response pairs. We then apply GRPO~\cite{shao2024deepseekmath} to optimize a response policy aligned with the shared and group-specific values encoded in the reward signals. Full details of the hyperparameter configuration are provided in Table~\ref{tab:gsm8k_hh_hyperparams}.

\paragraph{Results.} Tables~\ref{tab:HH_scores} and~\ref{tab: HH_winrates} summarize the performance of MaxMin- and SharedRep-RLHF on the HH dataset across two levels of minority representation. This is a setting in which the statistical minority is \textbf{not} the reward minority -- from Table~\ref{tab:HH_scores}, we observe that the majority has a negative average score across all experiments, while the minority has positive average scores. This experiment stands to test the performance of SharedRep-RLHF in settings which are not the most favorable in terms of showcasing the gain from SharedRep-RLHF. However, even in the less favorable setting, SharedRep-RLHF improves the mean minority score at both $1$\% and $10$\% proportions, reducing the gap to the gold preference score. For example, at $1$\% minority proportion, the minority score rises from $-0.038$ (MaxMin) to $0.017$ (SharedRep), approaching the gold score of $0.032$. Furthermore, SharedRep-RLHF maintains or improves majority performance in win rate terms. At $1$\% minority proportion, SharedRep-RLHF achieves a minority win rate of $34$\% (vs. $33$\% for MaxMin-RLHF) and a majority win rate of $43$\% (vs. $41$\%). Similar trends hold at $10$\% proportion. These results indicate that SharedRep-RLHF can better align with minority preferences without compromising -- and occasionally improving majority-aligned performance in practice, including settings in which the statistical minority is not the reward minority. However, unsurprisingly, the gains from SharedRep-RLHF in the setting are less pronounced compared to the previous settings. We further investigate variations in performance of SharedRep-RLHF with the hyperparameter $K$, which we plot in Figure~\ref{fig:ablation_hh}. We observe that $K=2$ has better performance at both $1$\% and $10$\% minority proportions in terms of the average minority score. In terms of win rate, the performance of SharedRep-RLHF is robust to variations in $K$. Consequently, we choose $K=2$ in reporting our results in Tables~\ref{tab:HH_scores} and~\ref{tab: HH_winrates}.


\subsection{SharedRep-RLHF Reward Model \& Compute}

\begin{figure}[t]
    \centering
    \includegraphics[width=0.8\linewidth]{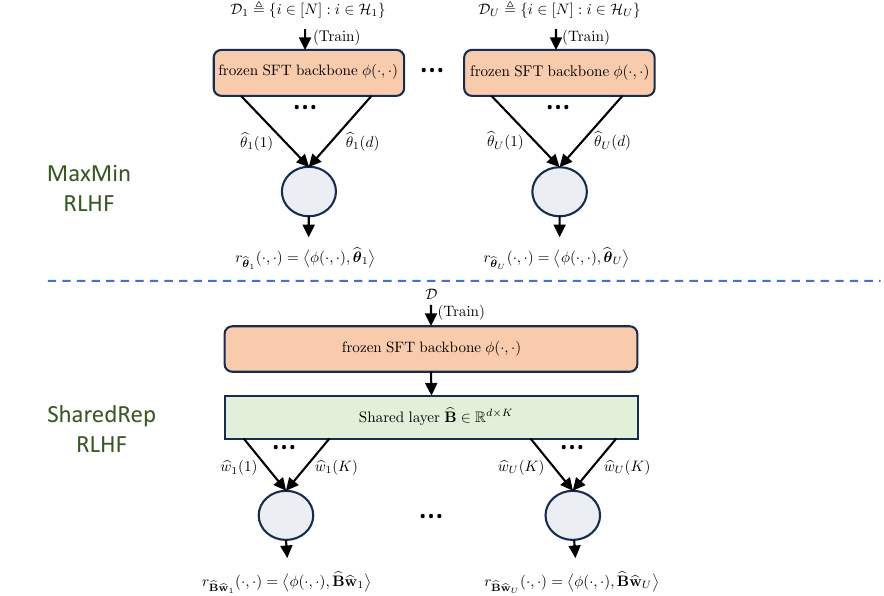}
    \caption{Reward model architectures of SharedRep-RLHF (comapred to MaxMin-RLHF)}
    \label{fig:architecture}
\end{figure}

We schematize the reward model architecture in Figure~\ref{fig:architecture}. For our experiments, we have used $8\times$A$6000$ GPUs with $49$Gb VRAMs to perform reward and RL training. Evaluations are performed on $2\times$RTX$3090$ GPUs with $24$Gb~VRAMs.



\end{document}